\documentclass[10pt,twocolumn,letterpaper]{article}

\usepackage{wacv}              

\usepackage{graphicx}
\usepackage{amsmath}
\usepackage{amssymb}
\usepackage{booktabs}

\usepackage{graphicx}
\usepackage{booktabs}

\usepackage{multirow}

\usepackage{subcaption}

\usepackage{array}

\usepackage{makecell}

\usepackage[accsupp]{axessibility} 

\usepackage[pagebackref,breaklinks,colorlinks]{hyperref}

\usepackage[capitalize]{cleveref}
\crefname{section}{Sec.}{Secs.}
\Crefname{section}{Section}{Sections}
\Crefname{table}{Table}{Tables}
\crefname{table}{Tab.}{Tabs.}

\def\wacvPaperID{1190} 
\def\confName{WACV}
\def\confYear{2025}

\begin{document}

\title{Improving Detail in Pluralistic Image Inpainting with Feature Dequantization}

\author{Kyungri Park\textsuperscript{\dag} and Woohwan Jung\\
Department of Applied Artificial Intelligence, Hanyang University\\
{\tt\small \{arten, whjung\}@hanyang.ac.kr}
}
\maketitle

\begin{abstract}
Pluralistic Image Inpainting (PII) offers multiple plausible solutions for restoring missing parts of images and has been successfully applied to various applications including image editing and object removal. 
Recently, VQGAN-based methods have been proposed and have shown that they significantly improve the structural integrity in the generated images.
Nevertheless, the state-of-the-art VQGAN-based model PUT faces a critical challenge: degradation of detail quality in output images due to feature quantization.
Feature quantization restricts the latent space and causes information loss, which negatively affects the detail quality essential for image inpainting.
To tackle the problem, we propose the FDM (Feature Dequantization Module) specifically designed to restore the detail quality of images by compensating for the information loss.
Furthermore, we develop an efficient training method for FDM which drastically reduces training costs.
We empirically demonstrate that our method significantly enhances the detail quality of the generated images with negligible training and inference overheads.
The code is available at \url{https://github.com/hyudsl/FDM}
\end{abstract}

\makeatletter
\renewcommand\@makefnmark{}
\makeatother
\footnotetext{\textsuperscript{\dag}Major in Bio Artificial Intelligence}

\newcommand{\vect}[1]{\mathbf{#1}}
\newcommand{\mask}{\vect{m}}
\newcommand{\defImg}{\img\in\mathbb{R}^{H\times W\times 3}}
\newcommand{\img}{\vect{x}}
\newcommand{\maskedImg}{\hat{\img}}
\newcommand{\defMaskedImg}{\maskedImg\in\mathbb{R}^{H\times W\times 3}}

\newcommand{\inpImg}{\img'}
\newcommand{\defInpImg}{\inpImg\in\mathbb{R}^{H\times W\times 3}}

\newcommand{\review}[1]{\vspace{-0.1in} \noindent\todo[color=lightgray,inline]{#1}}

\newcommand{\defFeat}{\feat\in\mathbb{R}^{\frac{H}{r}\times \frac{W}{r}\times C}}
\newcommand{\feat}{\mathbf{f}'}
\newcommand{\defMaskedFeat}{\maskedFeat\in\mathbb{R}^{\frac{H}{r}\times \frac{W}{r}\times C}}
\newcommand{\maskedFeat}{\hat{\mathbf{f}}}
\newcommand{\defInpFeat}{\inpFeat\in\mathbb{R}^{\frac{H}{r}\times \frac{W}{r}\times C}}
\newcommand{\inpFeat}{\mathbf{f}}
\newcommand{\defQuantFeat}{\quantFeat\in\mathbb{R}^{\frac{H}{r}\times \frac{W}{r}\times C}}
\newcommand{\quantFeat}{\mathbf{f}'_q}
\newcommand{\defQuantInpFeat}{\quantInpFeat\in\mathbb{R}^{\frac{H}{r}\times \frac{W}{r}\times C}}
\newcommand{\quantInpFeat}{\mathbf{f}_q}
\newcommand{\defDequantFeat}{\dequantFeat\in\mathbb{R}^{\frac{H}{r}\times \frac{W}{r}\times C}}
\newcommand{\dequantFeat}{\mathbf{f}'_{dq}}
\newcommand{\defDequantInpFeat}{\dequantInpFeat\in\mathbb{R}^{\frac{H}{r}\times \frac{W}{r}\times C}}
\newcommand{\dequantInpFeat}{\mathbf{f}_{dq}}
\newcommand{\defQuantError}{\quantError\in\mathbb{R}^{\frac{H}{r}\times \frac{W}{r}\times C}}
\newcommand{\quantError}{\mathbf{f}'_{qe}}
\newcommand{\defInpQuantError}{\inpQuantError\in\mathbb{R}^{\frac{H}{r}\times \frac{W}{r}\times C}}
\newcommand{\inpQuantError}{\mathbf{f}_{qe}}
\newcommand{\defPredQuantError}{\predQuantError\in\mathbb{R}^{\frac{H}{r}\times \frac{W}{r}\times C}}
\newcommand{\predQuantError}{\tilde{\mathbf{f}}_{qe}}

\newcommand{\codebookSize}{\mathcal{N}}
\newcommand{\samplingNum}{\mathcal{K}}

\newcommand{\ourmodule}{FDM\xspace}
\newcommand{\putModel}{PUT\xspace}
\newcommand{\mat}{MAT\xspace}
\newcommand{\ict}{ICT\xspace}
\newcommand{\ldm}{LDM\xspace}
\newcommand{\lama}{LaMa\xspace}
\newcommand{\df}{DeepFill v2\xspace}
\newcommand{\cmt}{CMT\xspace}
\newcommand{\vqgan}{VQGAN\xspace}
\newcommand{\ourPUT}{Ours (\putModel+\ourmodule)\xspace}
\newcommand{\ourVQGAN}{Ours (\vqgan+\ourmodule)\xspace}

\newcommand{\minisection}[1]{%
\vspace{0.04in}
    \noindent \textbf{#1}.\xspace%
}

\newcommand{\red}[1]{\textcolor{red}{#1}}
\newcommand{\blue}[1]{\textcolor{blue}{#1}}

\newcommand{\memo}[1]{\textcolor{red}{#1}}

\newcommand{\todomemo}[1]{\todo{#1}}

\section{Introduction}
\label{sec:intro}

Pluralistic Image Inpainting (PII), which offers multiple plausible solutions for missing parts of images, has gained attention for enhancing user satisfaction in real-world applications.
By providing a variety of candidates, PII not only increases user engagement but also ensures that the final results align closely with individual preferences.
Consequently, PII has been applied to various real-world tasks, including image editing \cite{Jo_2019_ICCV} and object removal \cite{shetty2018adversarial, din2020novel}.

Recent studies \cite{wan2021high, liu2022reduce} have demonstrated that sampling values for masked regions in the codebook can generate more diverse and well-structured images.
\ict\cite{wan2021high} sampling the RGB value of the pixel-level codebook (a set of RGB values) for each masked pixel in the image.
However, to reduce the sampling cost, \ict downsamples the input image, leading to information loss.
To address this issue, \putModel \cite{liu2022reduce} encodes each patch into features instead of downsampling.
\putModel samples features from a patch-level codebook (a set of features) for each masked patch, and the decoder reconstructs the output image using a quantized feature map consisting of the selected features.

\begin{figure}[tb]
  \centering
  \begin{subfigure}[t]{0.32\linewidth}
    \includegraphics[width=\textwidth]{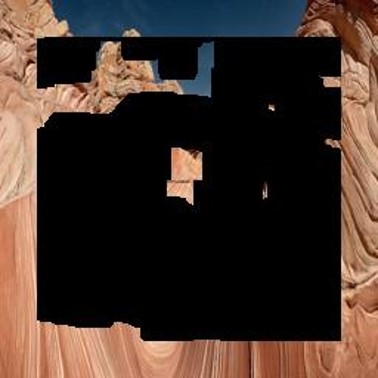}
    \caption*{Masked Image}
  \end{subfigure}
  \begin{subfigure}[t]{0.32\linewidth}
    \includegraphics[width=\textwidth]{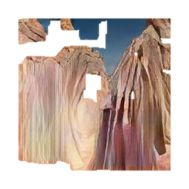}
    \caption*{Generated Area}
  \end{subfigure}
  \begin{subfigure}[t]{0.32\linewidth}
    \includegraphics[width=\textwidth]{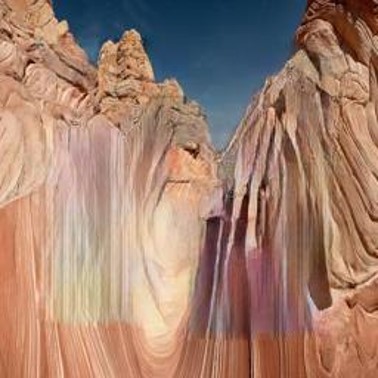}
    \caption*{Inpainted Image}
  \end{subfigure}
  \caption{An example of the visible boundary between the generated area and the masked image caused by feature quantization.
  Although the generated area in the center image appears plausible and realistic, a slight color mismatch at the boundary makes it noticeable when combined with the masked image (on the right).
  }
  \label{fig:problem_inpainting}
\end{figure}

However, feature quantization with the codebook still leads to information loss by restricting the decoder's input space to a discrete feature space.
This information loss can degrade the detail quality in output images.
In image inpainting, degraded detail quality may result in mismatches between the texture and color of the generated area and the surrounding background.
Such inconsistencies can create noticeable boundaries between the generated area and the background, causing the final image to appear unnatural.
For instance, while the image in the center of Figure~\ref{fig:problem_inpainting} appears realistic, the image on the right shows clear visibility of the masked area due to color mismatches when merging with the background.

To address this problem, we introduce \ourmodule (Feature Dequantization Module), designed to estimate the gap (error) between the ideal features and the quantized features.
\ourmodule compensates for this gap by adding the estimated corrections to the quantized features, significantly improving the detail quality of the generated images.
However, a straightforward end-to-end training of \ourmodule is infeasible due to the iterative codebook sampling in VQGAN.
Therefore, we propose an efficient method to train \ourmodule that reduces training costs by over 99\% based on our estimation.
Moreover, the inference cost of \ourmodule is negligible, as feature dequantization is performed only once per input image.

To evaluate the effectiveness of FDM, we conducted comparative experiments with state-of-the-art PII models.
The results of our experiments indicated that applying FDM enhances the details and structural consistency while preserving diversity.
Furthermore, we applied FDM to various image generation tasks beyond inpainting.
These experiments consistently demonstrated improved performance, highlighting the effectiveness of FDM across a wide range of \vqgan-based image generation tasks.

The contributions of this paper are as follows:
\begin{itemize}
    \item We identify the information loss in the state-of-the-art pluralistic image inpainting model, \putModel.
    \item We propose the \ourmodule which significantly improves the detail quality and consistency of the generated images by compensating for the information loss.
    \item We develop an efficient training method for \ourmodule which drastically reduces training costs without sampling procedure.
\end{itemize}
\section{Related Work}
\label{sec:related_work}

\minisection{Pluralistic image inpainting methods}
PII can generate multiple results for each input, unlike conventional inpainting methods that typically produce a single result.
VAE\cite{kingma2014auto}-based methods \cite{zheng2019pluralistic, liu2021pd} have been proposed to enable diverse image generation.
PIC\cite{zheng2019pluralistic} encodes masked inputs into a Gaussian distribution, generating diverse images via latent vector sampling.
PD-GAN\cite{liu2021pd} combines prior inpainted images and SPADE\cite{park2019semantic} to enhance diversity, with latent vector decoding conditioned on deterministic inpainting results.
Recent Transformer-based methods\cite{wan2021high, liu2022reduce, li2022mat} outperform VAE-based approaches.
\ict \cite{wan2021high} generates low-resolution prior image through the Transformer and then up-samples the prior image using a CNN to generate the final results.
\putModel\cite{liu2022reduce} utilizes the \vqgan architecture\cite{esser2021taming} and effectively generates diverse results by reducing input information loss through a patch-based encoder.
However, it has limited representational capacity due to the use of a discrete codebook, which can lead to distorted structures and color discrepancies.
To address this issue, we propose FDM, which increases the representational capacity through feature dequantization.

\minisection{Image generation methods with \vqgan}
\vqgan, based on VQ-VAE\cite{van2017neural}, is a method designed for high-resolution image generation.
While \vqgan and its variants \cite{huang2023not, cao2023efficient, huang2023towards} efficiently generate images using a codebook, the discrete latent space of the codebook imposes limitations on its representational capacity compared to conventional VAE-based image generation methods which typically operate in a continuous latent space.
Several methods \cite{lee2022draft, lee2022autoregressive, you2022locally, lin2023catch} have been proposed to enhance the representational capacity of codebooks in image generation.
The RQ-VAE \cite{lee2022autoregressive} introduces the use of multiple codes to represent the latent vector through a residual quantizer.
In contrast to employing multiple codebooks, the RQ-VAE efficiently enhances representational capacity by utilizing a single codebook while employing multiple codes.
RQ-VAE needs additional patch sampler for multiple codes.
In contrast, our proposed method can easily enhance performance by adding a very small module without altering the overall structure.
FA-VAE \cite{lin2023catch} aims to recover missing details that occur during feature quantization in the frequency domain.
It achieves this by learning to match the frequency of the feature map output during the decoding process with that of the feature map during the encoding process, effectively restoring missing details.
In contrast to these previous methods, we propose FDM to enhance the representational capacity for pluralistic image inpainting.
\section{Proposed Method}
\label{sec:method}

Our proposed method aims to enhance the details of inpainted images using the Feature Dequantization Module (\ourmodule).
We begin with an overview of our method, followed by a training strategy for \ourmodule, and finally provide the entire training procedure.

\begin{figure*}[tb]
    \centering
    \includegraphics[width=0.85\textwidth]{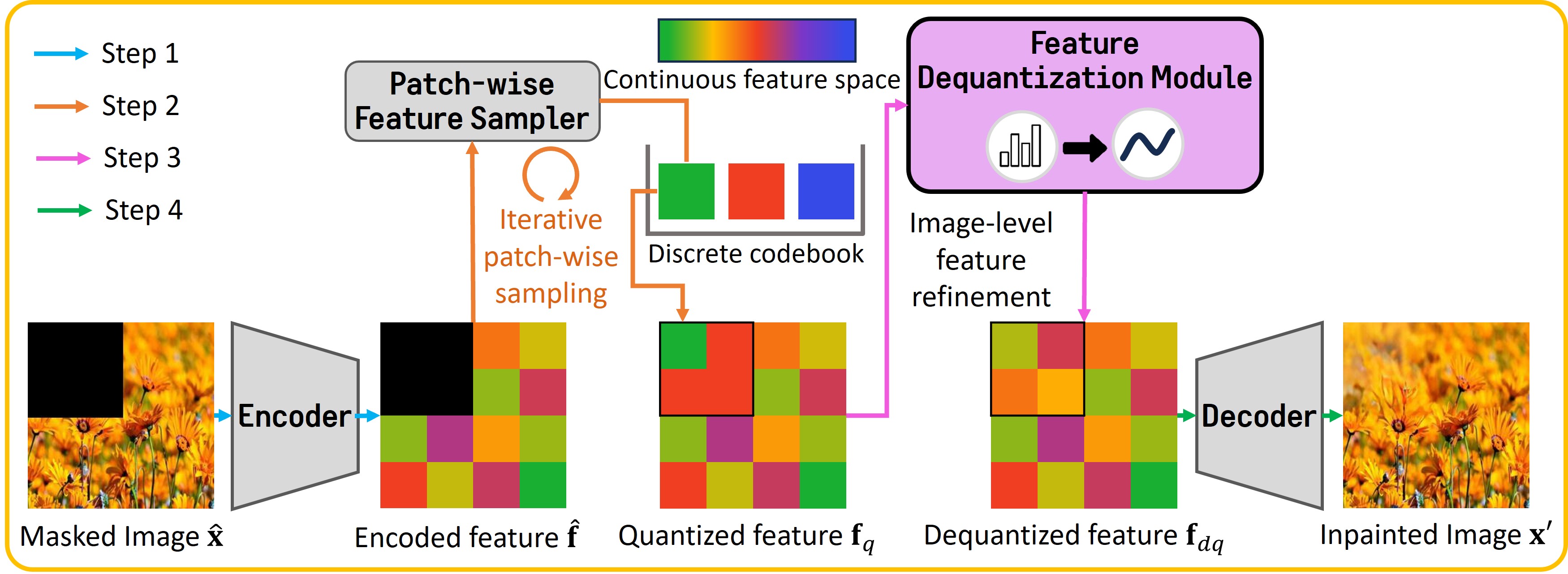}
    \caption{An overview of the proposed method. 
    Quantized features are limited to distinct points, represented by green, red, and blue. 
    However, through the proposed \ourmodule module, dequantization expands the representation to a continuous space.}
    \label{fig:overview}
\end{figure*}

\subsection{Overview}
\label{sec:architecture}
Let $\defImg$ be an image and $\mask \in\{0,1\}^{H\times W}$ be a binary mask, where $H$ and $W$ represent the height and width of the image, respectively.
The masked image is represented as $\defMaskedImg$ where
$\maskedImg = \img \otimes \mask$ and $\otimes$ denotes element-wise multiplication.
Pixels with a value of 0 in $\maskedImg$ will be inpainted.
Our goal is to generate diverse images that contain content similar to the original image $\img$, starting from the masked image $\maskedImg$.
An overview of our proposed inpainting procedure consists of the following four steps as illustrated in Figure~\ref{fig:overview}: Encoding, feature sampling, feature dequantization, and decoding.
It should be noted that, except for Step 3, our approach aligns with PUT \cite{liu2022reduce}, and additionally removing Step 2 yields a variational auto-encoder (VAE). 
We introduce the dequantization (Step 3), to address the issue of information loss occurred by the patch-wise feature sampling (Step 2).

\minisection{Step 1: Encoding}
We first partition the masked image $\maskedImg$ into $\frac{H}{r}\times \frac{W}{r}$ non-overlapping patches, each of size $r\times r$. 
For each patch, the encoder generates a $C$-dimensional feature vector.
The collection of this feature vectors from a feature map, represented as $\defMaskedFeat$.
Following \cite{liu2022reduce}, we set the patch size $r$ to 8 and the dimensionality $C$ to 256.
We provide detailed information about the structure of the encoder-decoder in the supplementary material.

\minisection{Step 2: Patch-wise feature sampling}
The feature-sampler takes the encoded feature map $\maskedFeat$ as input and outputs a quantized feature $\quantInpFeat$.
In this step, the feature vector of each masked patch is replaced by a vector sampled from a codebook which is a set of $\codebookSize$ possible feature vectors.
To produce diverse results, each patch is sampled autoregressively using Gibbs sampling from a probability distribution predicted by a patch-wise feature-sampler.
Based on the learned distribution, the feature-sampler can sample from among the most suitable $\samplingNum$ vectors, where $\samplingNum$ represents a hyper-parameter that controls the diversity of sampling.
Since the codebook contains a limited number of possible latent vectors, the feature-sampler can easily learn distribution of latent vector in each patch.
The masked features are replaced with features from the codebook, which are sampled by the feature-sampler.

However, the sampling with codebook induces the problem of feature quantization.
The feature map generated by the feature-sampler is quantized to the codebook’s feature vectors.
Feature quantization restricts the decoder to access only limited points, causing information loss.
Consequently, this information loss reduces the representational capability and degrades the detail quality of the output image.
In image inpainting, where consistency with the background is essential, detail degradation can lead to mismatched colors and texture with the background, resulting in lower image quality.
While increasing size of the codebook improves approximation accuracy of the continuous feature space, it does not completely resolve the issue of information loss and increases sampling difficulty.

\minisection{Step 3: Feature dequantization}
To address information loss caused by feature quantization, we introduce a simple solution called \ourmodule (Feature Dequantization Module) which involves dequantizing the sampled features.

Quantization error is defined as the difference between the original continuous feature and its quantized representation.
\ourmodule compensates for the error by adding the estimated quantization error to the quantized features.
Let $\defInpFeat$ be the ideal continuous feature related to the quantized feature $\quantInpFeat$.
The quantization error can be represented as $\inpQuantError=\inpFeat-\quantInpFeat$. 
\ourmodule predicts $\inpQuantError$ and adds it to $\quantInpFeat$ to achieve dequantization.

\ourmodule takes $\quantInpFeat$ and the downsampled mask $\mask_d\in\mathbb{R}^{\frac{H}{r}\times\frac{W}{r}\times C}$ as inputs to predict the quantization error:
\begin{align*}
\predQuantError=\mathcal{F}_\theta([\quantInpFeat,\mask_d]) 
\end{align*}
where $\predQuantError$ is predicted quantization error by the estimation function $\mathcal{F}_\theta$. Note that the feature of unmasked patch in $\quantInpFeat$ is the same as $\maskedFeat$ because the feature-sampler only replaces the features of masked patches.
The estimation function $\mathcal{F}_\theta$ is a simple network composed of 8 residual blocks \cite{he2016deep}, each consisting of a 3x3 convolutional layer followed by a ReLU activation function and a 1x1 convolutional layer.
After the 1x1 convolutional layer, the output is added to the input of the block and passed through a ReLU activation function.
Residual blocks were chosen for their ability to learn while preserving input information, ensuring structural similarity with quantized features even after dequantization.

The predicted quantization error $\predQuantError$ is added to $\quantInpFeat$ for dequantization:
\begin{align*}
\dequantInpFeat=\quantInpFeat+\predQuantError\otimes (1-\mask_d)
\end{align*}

where $\otimes$ represents the element-wise multiplication operation, and $\dequantInpFeat$ is the predicted dequantized feature.
It is worth noting that the feature dequantization is performed done only once for an input image. 
By doing this, we can minimize the computational overhead.
Additionally, \ourmodule is applied after all patches have been sampled.
It complements the features of each patch across the entire image, thereby enhancing the consistency of image.

\minisection{Step 4: Decoding}
The decoder takes the dequantized feature $\dequantInpFeat$ as input and generates an inpainted image $\inpImg$:
$$\inpImg = Decoder(\dequantInpFeat)$$ where the decoder is a convolutional neural network.

\subsection{Training the Feature Dequantization Module}
\label{sec:training_fdm}
The full model training with the reconstruction loss is a straightforward method to train \ourmodule.
However, it incurs substantial costs particularly due to the iterative sampling. 
Thus, we propose an efficient method to train \ourmodule without the feature-sampler and the decoder to minimize the training cost.
Additionally, this method prevents the potential catastrophic forgetting that may occur when the decoder is jointly trained with a randomly initialized \ourmodule, offering a cost-effective and reliable alternative to the conventional, more expensive training process.

\minisection{Quantization error prediction}
Recall that \ourmodule predicts the quantization error $\inpQuantError$ by the following equation: $\predQuantError=\mathcal{F}_\theta([\quantInpFeat,\mask_d])$. 
Thus, we can directly train \ourmodule to predict the quantization error  $\inpQuantError$ without the decoder
with the following quantization error prediction loss $L_{qe}$:
\begin{equation}
\label{eq:loss_qe_origin}
\mathcal{L}_{qe} = |\inpQuantError-\mathcal{F}_\theta([\quantInpFeat,\mask_d])|.   
\end{equation}

However, the training with the above loss involve the iterative patch-wise feature sampling process to generate $\quantInpFeat$.
It causes a significant overhead in the training process since the number of iterations is equal to the number of patches $\frac{H}{r}\cdot\frac{W}{r}$(1024 in our experiments).
To avoid the excessive training cost, we propose a method to train without the feature-sampler.

\minisection{Training without the feature-sampler}
The main idea to eliminate the feature-sampler from the training process is to utilize unmasked images.
Firstly, we encode the unmasked image $\img$ using a frozen encoder to obtain the encoded unmasked feature $\defFeat$.
Then, the feature-sampler quantizes the feature $\feat$ to obtain $\defQuantFeat$.
Finally, the target value $\quantError$ is calculated as $\feat\otimes \mask -\quantFeat\otimes (1-\mask_d)$.
In other words, we approximate $\mathcal{L}_{qe}$ in Equation~\eqref{eq:loss_qe_origin} using this formulation:
\begin{equation*}
\label{eq:loss_qe_approx}
\mathcal{L}_{qe} \approx |\quantError-\mathcal{F}_\theta([\quantFeat,\mask_d])|   
\end{equation*}
where $\quantError = \feat\otimes \mask -\quantFeat\otimes (1-\mask_d)$.

Given the identical sampling setting used in evaluation, one training iteration with the patch sampling consumes around 387 seconds with 16 batches, making training cumbersome.
Our modification reduces the training time to approximately 2.1 second per iteration, making \ourmodule training 184 times faster compared to the training with patch-sampling method.

\begin{figure}
    \centering
    \includegraphics[width=0.9\linewidth]{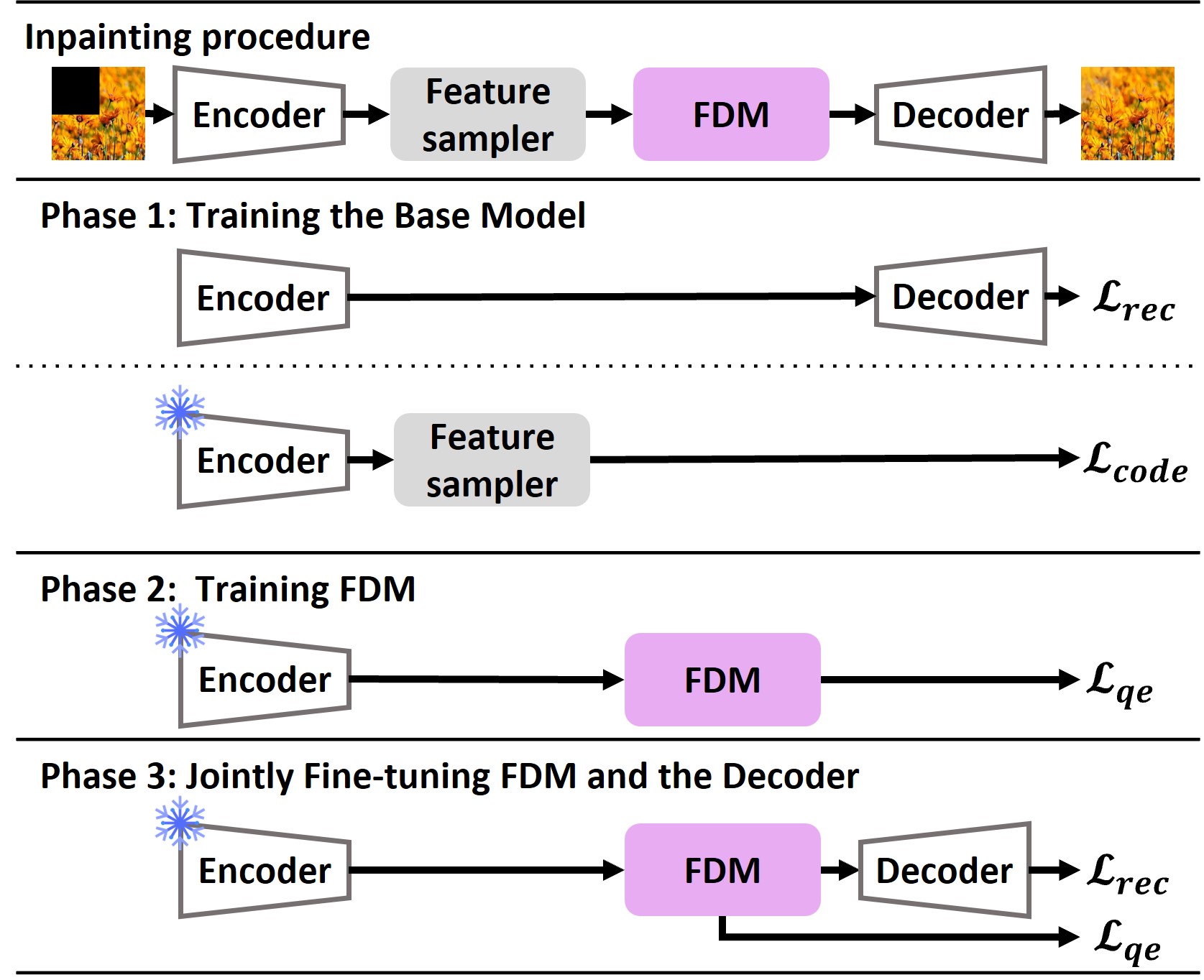}
    \caption{An overview of the training procedure.}
    \label{fig:tr_procedure}
\end{figure}

\subsection{Training Procedure}
\label{sec:training_procedure}
Figure~\ref{fig:tr_procedure} shows the entire training procedure of our proposed method.
First, we train the encoder-decoder and the feature-sampler as in \cite{liu2022reduce}.
Next, we train \ourmodule as presented in \ref{sec:training_fdm}.
Finally, we jointly fine-tune \ourmodule and the decoder.

\minisection{Phase 1: Training the base model} We train the base model \putModel which consists of encoder-decoder and a patch-wise feature-sampler.
To train the encoder-decoder, we employ the image reconstruction loss $\mathcal{L}_{rec}$ between the target image and the reconstructed image.
It is a weighted sum of L1 loss $\mathcal{L}_{l1}$, gradient loss $\mathcal{L}_{G}$, adversarial loss \cite{goodfellow2014generative} $\mathcal{L}_{A}$, perceptual loss \cite{johnson2016perceptual} $\mathcal{L}_{P}$, and style loss \cite{gatys2016image} $\mathcal{L}_{S}$:
\begin{align}
\label{eq:loss_rec}
\mathcal{L}_{rec} = \mathcal{L}_{l1}+\lambda_{G}\mathcal{L}_{G}+\lambda_{A}\mathcal{L}_{A}+\lambda_{P}\mathcal{L}_{P}+\lambda_{S}\mathcal{L}_{S}.
\end{align}
We set $\lambda_{G} = 5$, $\lambda_{A} = 0.1$, $\lambda_{P} = 0.1$ and $\lambda_{S} = 250$ following \cite{liu2022reduce}.
For training the feature-sampler, we utilize the code classification loss $\mathcal{L}_{code}$.
This loss measures the cross-entropy between the predicted code class probability of each patch and the target code class, where the code class indicates the latent vector of the codebook.

In phase 1, the encoder-decoder is initially trained using $\mathcal{L}_{rec}$ for image reconstruction. Then, with the encoder frozen, the feature-sampler is trained using $\mathcal{L}_{code}$ to learn the distribution of latent vectors per patch. While the trained encoder-decoder and feature-sampler alone can construct an inpainting model, we further enhance performance by integrating the \ourmodule, into the model.

\minisection{Phase 2: Training \ourmodule}
In Phase 2, the feature dequantization module is trained using $\mathcal{L}_{qe}$ as described in Section~\ref{sec:training_fdm}, with the encoder frozen during this process.
We freeze the encoder to prevent them from forgetting previously learned features.

\minisection{Phase 3: Jointly fine-tuning \ourmodule and the decoder}
After the initial training \ourmodule, we jointly fine-tune \ourmodule and the decoder which are separately trained before this phase.
The loss function for the fine-tuning $\mathcal{L}_{rec}$ is as follows:
\begin{align}
\mathcal{L}_{tuning} = \lambda_{qe}\mathcal{L}_{qe} + \lambda_{rec}\mathcal{L}_{rec}
\end{align}
where $\lambda_{qe}=1$, $\lambda_{rec}=1$, and $\mathcal{L}_{rec}$ remains the same as in Equation~\eqref{eq:loss_rec}.
Since the input of the decoder is modified by \ourmodule, we fine-tune the decoder.
To ensure that \ourmodule does not forget its ability to predict quantization error, we continue to use $\mathcal{L}_{qe}$ during training.
\section{Experiments}
\label{sec:experiments}

\subsection{Datasets and Settings}
\label{sec:dataset}

\minisection{Datasets}
The evaluation is conducted at 256 × 256 resolution on two datasets: Places \cite{zhou2017places} and Paris Street View Dataset \cite{doersch2012makes}.
Irregular masks provided by PConv \cite{liu2018image} are used for both training and testing.
In the experimental results, mask ratios between 0.2 and 0.4 are referred to as small masks, while those between 0.4 and 0.6 are labeled as large masks.
Following \cite{wan2021high}, we only utilized a subset of Places for our experiments, while keeping the training and test splits consistent with the original dataset.
We keep 237,777 images for training and reserve 800 images for testing.
In the Paris Street View Dataset, for consistent evaluation, we reorganized the existing split to use 14,000 images for training and 1,000 images for testing.

\minisection{Metric}
Evaluation is conducted using Fr\'echet Inception Distance (FID) \cite{heusel2017gans} and Learned Perceptual Image Patch Similarity (LPIPS) \cite{zhang2018unreasonable}. 
We selected FID and LPIPS as evaluation metrics because they closely resemble human perceptual capabilities.
In contrast, PSNR, SSIM \cite{wang2004image}, and MAE are not well-aligned with human perceptual capabilities \cite{ledig2017photo, sajjadi2017enhancenet} due to their pixel-wise calculations.
Therefore, they are not suitable for evaluating pluralistic inpainting.
However, FID and LPIPS evaluate images in the feature space, allowing for judgments that are much closer to human perception, making them more suitable for pluralistic inpainting.
Thus, we provided result of PSNR, SSIM, and MAE in the supplementary materials.
For evaluation, we use only one generated result per input.

We evaluate diversity score using LPIPS similar to \cite{wan2021high}.
Unlike evaluating inpainting performance, when measuring diversity, LPIPS is computed using only the generated images.
First, we generate $N$ paired pluralistic inpainted images using the same mask.
Then, the mean LPIPS score for each pair is used as the diversity score.
For evaluation, we generated $N=5$ pairs for each image.

\minisection{Compared methods}
We compare the proposed method with the following state-of-the-art pluralistic inpainting approaches: \ict\cite{wan2021high}, \mat\cite{li2022mat}, \ldm\cite{rombach2022high} and \putModel\cite{liu2022reduce}.
We evaluate using the provided pre-trained models.
In cases where pre-trained models are not available, we train the models using the code and settings provided by the authors.

\minisection{Experimental settings}
The detailed structures of the encoder-decoder and feature-sampler follow the \putModel architecture and use the same model size for both datasets.
We set the patch size $r$ to 8, resulting in 1024 patches for a $256\times256$ resolution image.
The number of latent vectors in the codebook $\codebookSize$ is set to 512.
We set $\samplingNum=50$ for pluralistic results in \ict, \putModel, and FDM.
Training of FDM utilizes the same settings as when training the encoder-decoder.
The learning rate is warmed up from 0 to 2e-4 in the first epoch and then decayed with a cosine scheduler.
FDM and decoder are optimized with Adam \cite{kingma2014adam} $(\beta_1 = 0, \beta_2 = 0.9)$.
The FDM and encoder-decoder are trained for 100 epochs, while the feature sampler is trained for 300 epochs. 
Training stops if there is no improvement in the baseline metric—validation loss for encoder-decoder and FDM, and classification accuracy for the feature sampler—over 10 consecutive epochs.
All models are trained on a machine with six GeForce RTX 2080 Ti GPUs.

\subsection{Main Results}
\label{sec:main_results}

\begin{figure*}[tb]
  \centering
  \begin{subfigure}[t]{0.12\linewidth}
    \includegraphics[width=\textwidth]{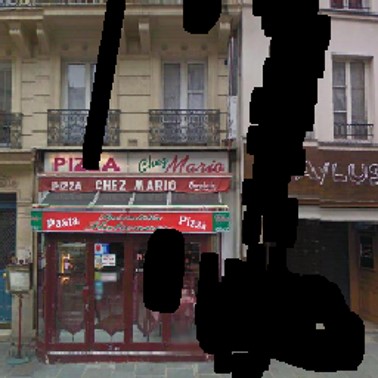} \\
    \includegraphics[width=\textwidth]{figure/detail/1_masked.jpg}
    \caption*{Input}
  \end{subfigure}
  \begin{subfigure}[t]{0.12\linewidth}
    \includegraphics[width=\textwidth]{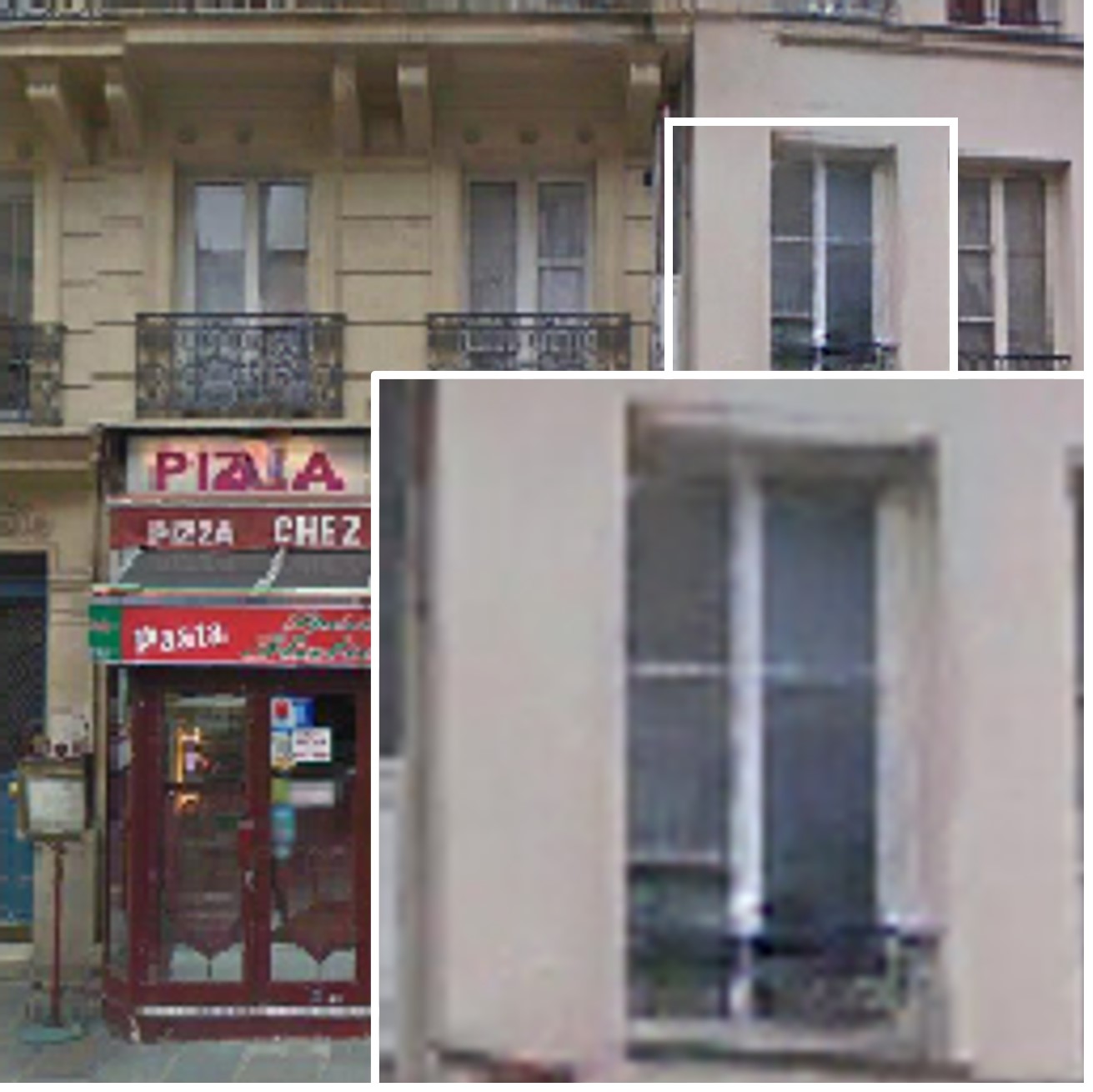} \\
    \includegraphics[width=\textwidth]{figure/detail/1_PUT.jpg}
    \caption*{PUT}
  \end{subfigure}
  \begin{subfigure}[t]{0.12\linewidth}
    \includegraphics[width=\textwidth]{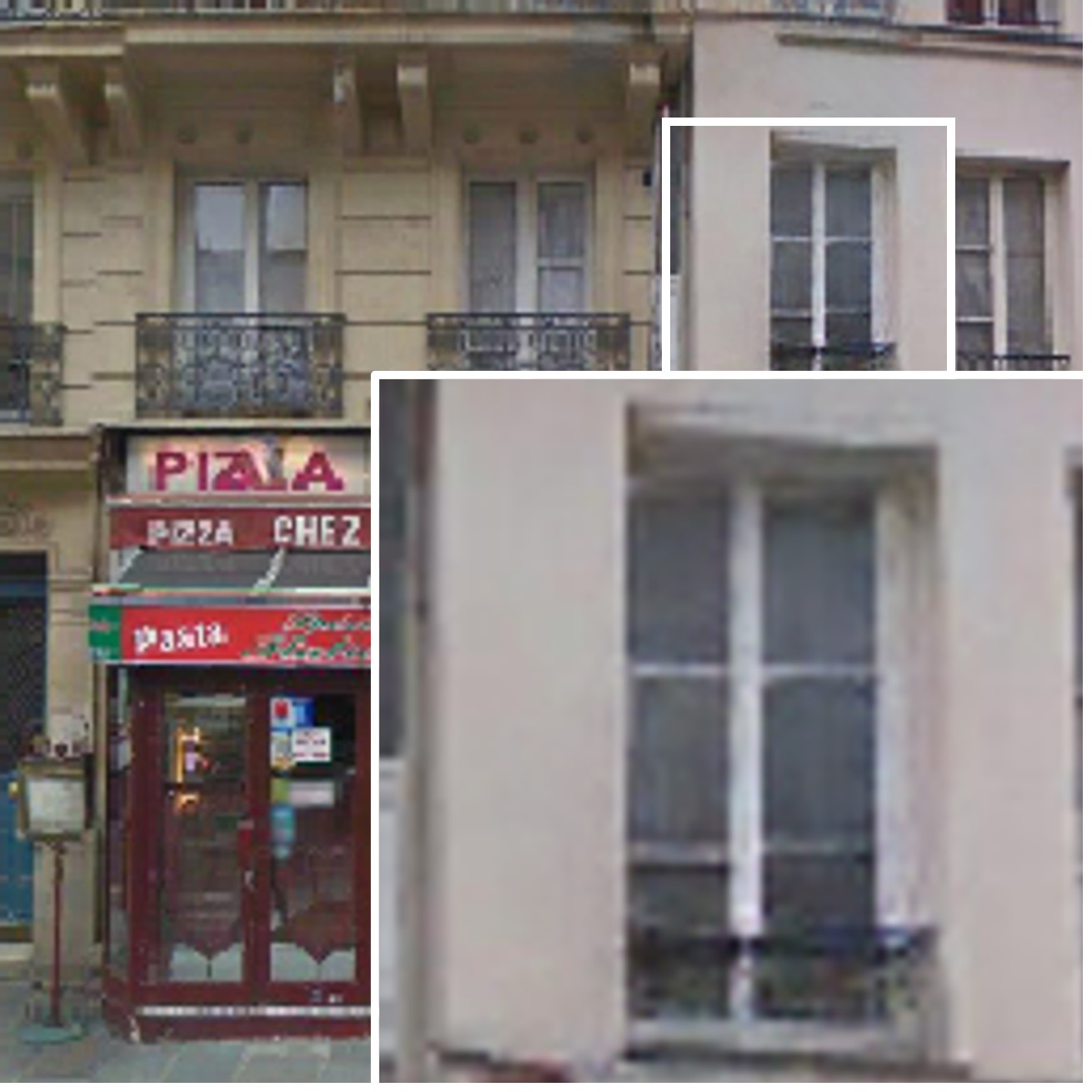} \\
    \includegraphics[width=\textwidth]{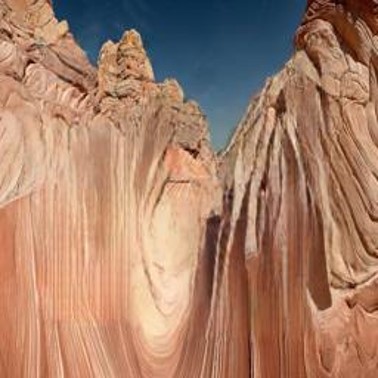}
    \caption*{Ours}
  \end{subfigure}
  \begin{subfigure}[t]{0.12\linewidth}
     \includegraphics[width=\textwidth]{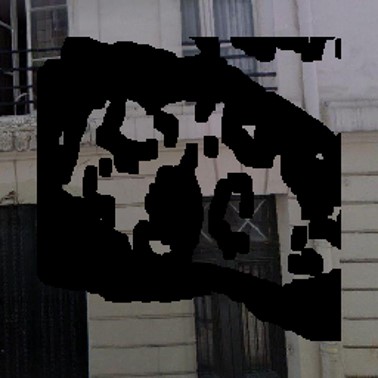} \\
     \includegraphics[width=\textwidth]{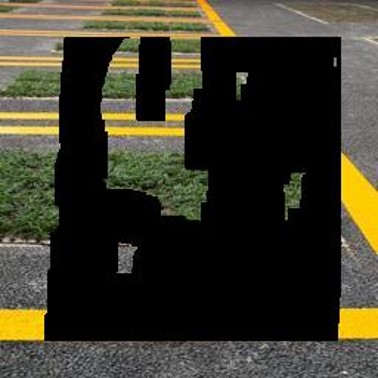}
    \caption*{Input}
  \end{subfigure}
  \begin{subfigure}[t]{0.12\linewidth}
    \includegraphics[width=\textwidth]{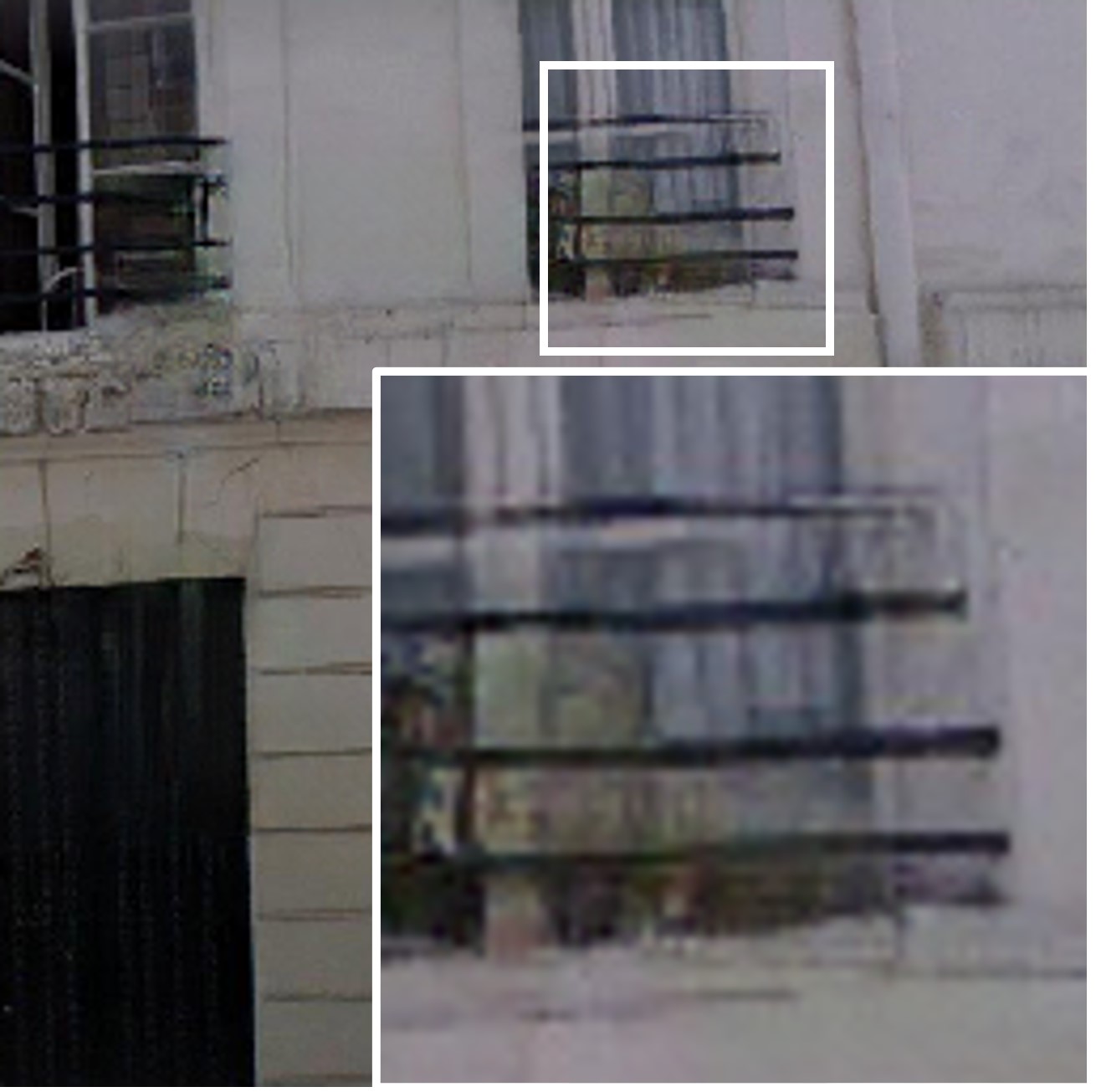} \\
    \includegraphics[width=\textwidth]{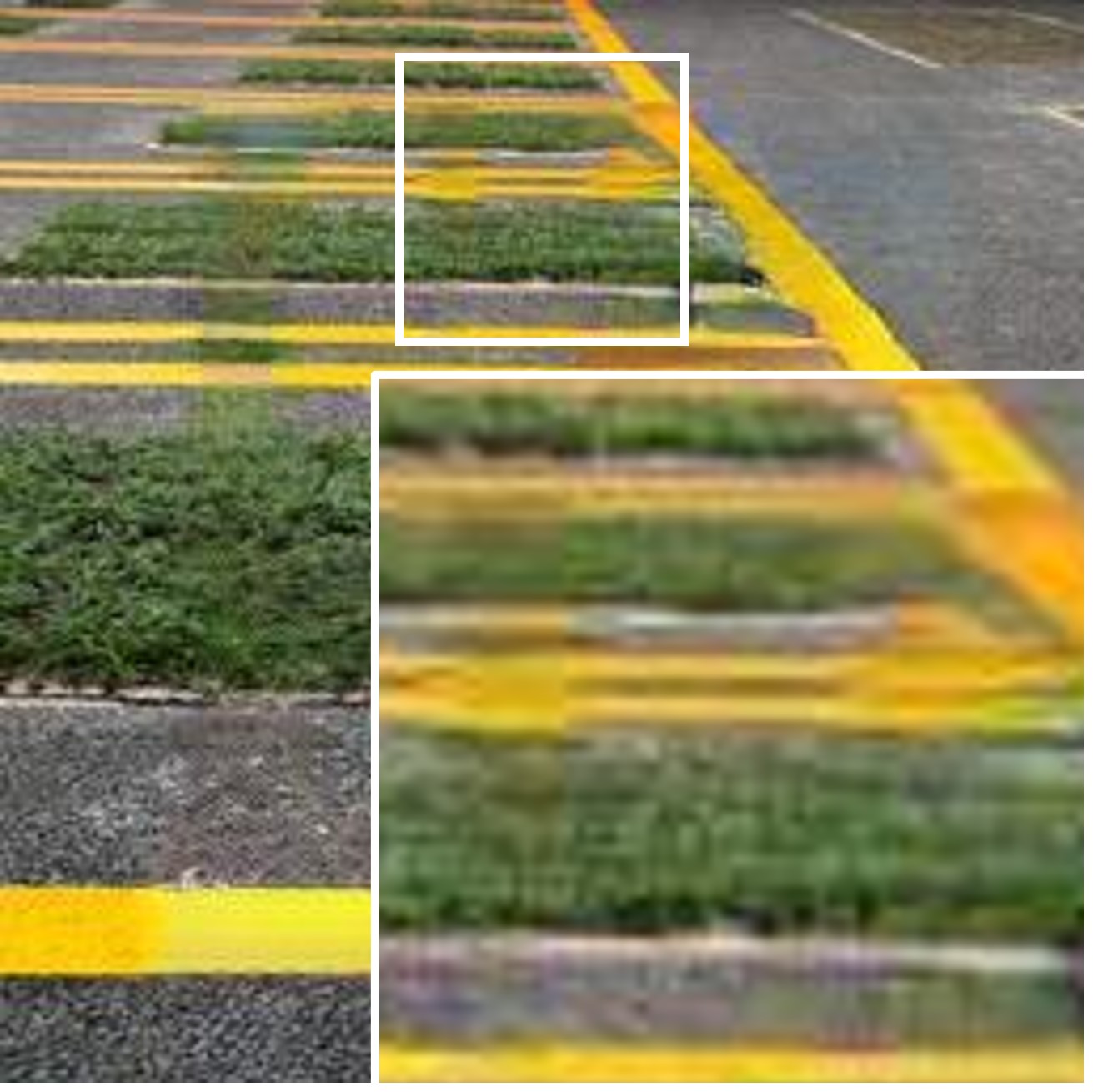}
    \caption*{PUT}
  \end{subfigure}
  \begin{subfigure}[t]{0.12\linewidth}
    \includegraphics[width=\textwidth]{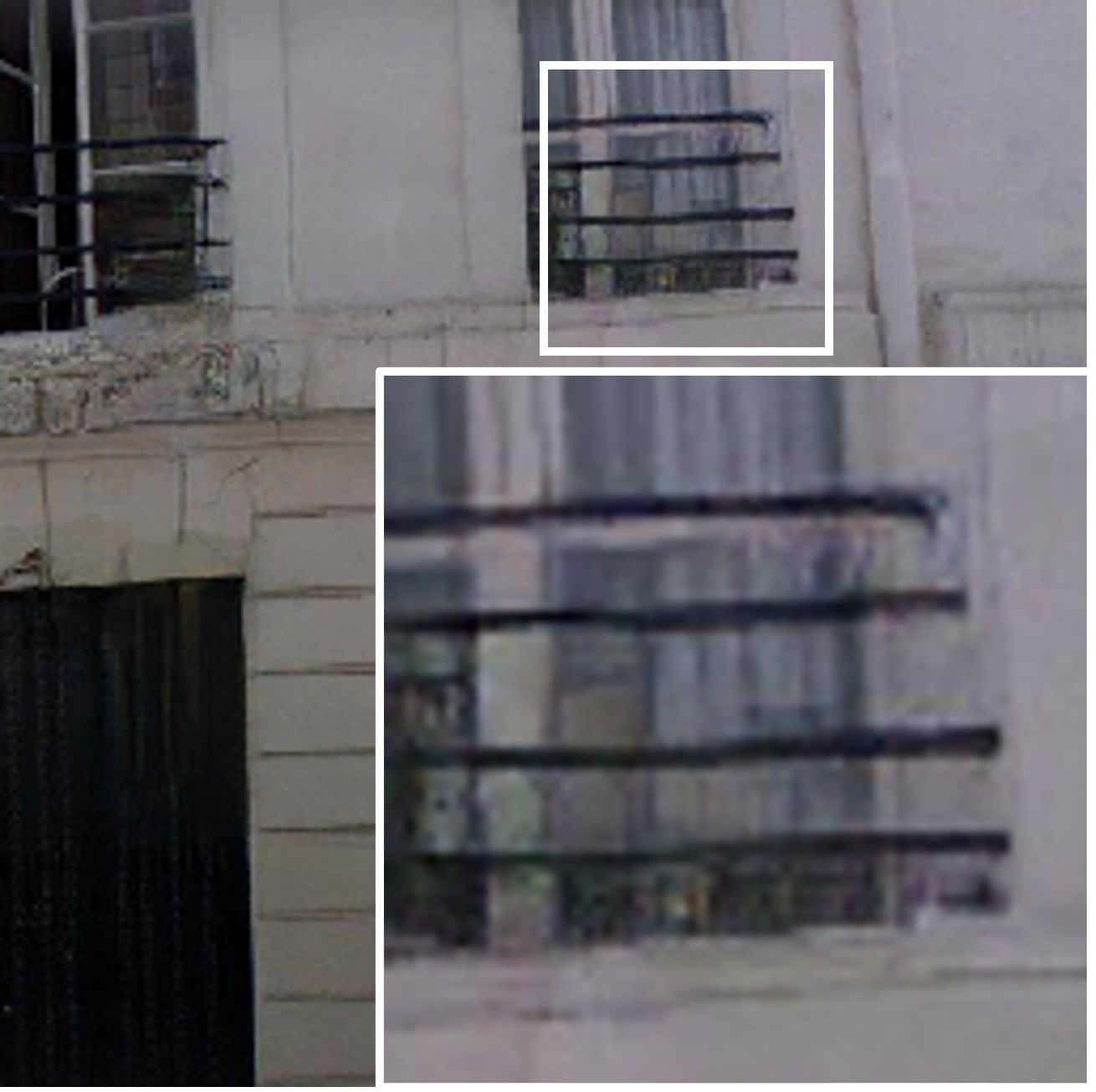} \\
    \includegraphics[width=\textwidth]{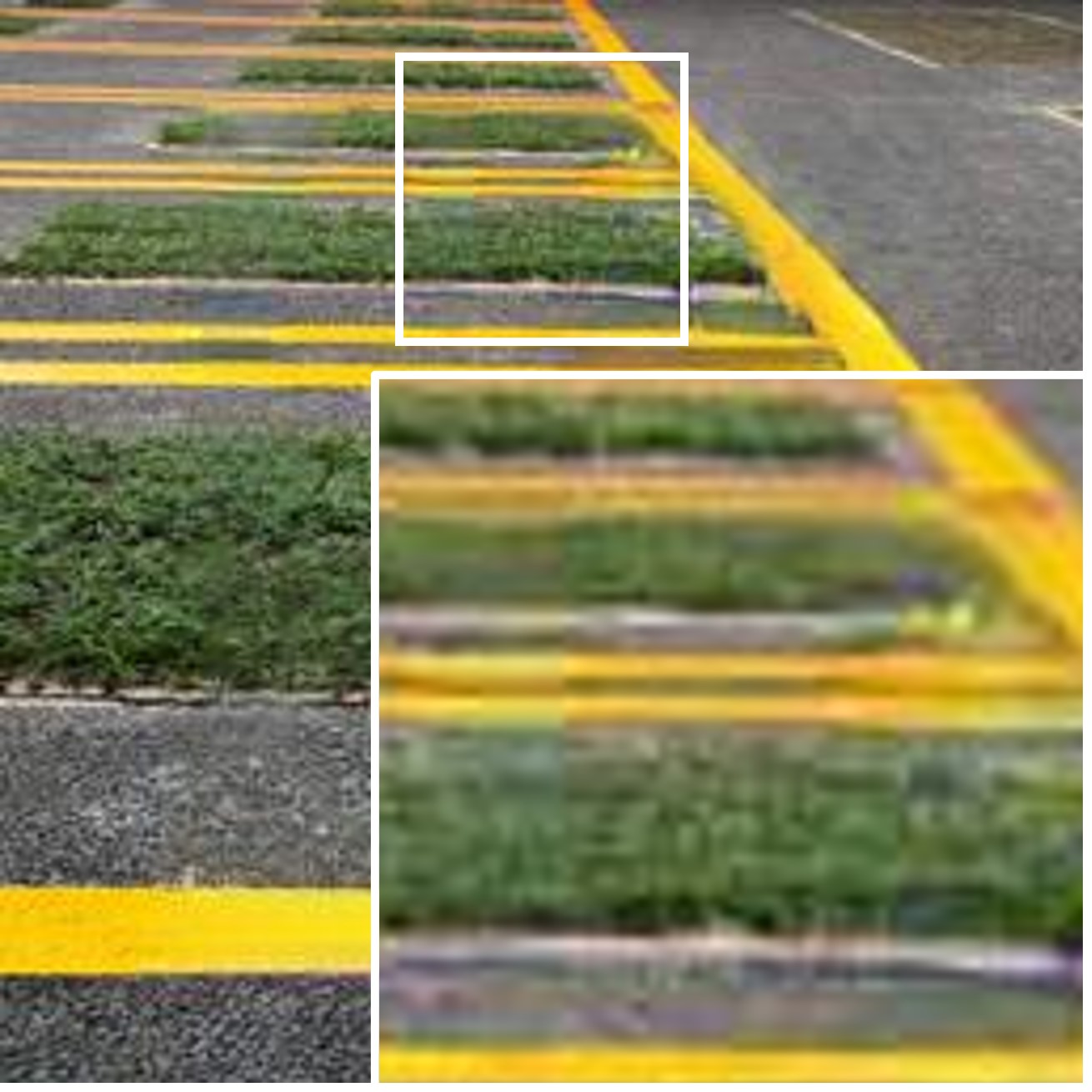}
    \caption*{Ours}
  \end{subfigure}
  \caption{Detail comparison between proposed method and PUT. More results are presented in the supplementary material.}
  \label{fig:detail_inp_result}
\end{figure*}

\begin{figure*}[tb]
  \centering
  \begin{subfigure}[t]{0.12\linewidth}
    \includegraphics[width=\textwidth]{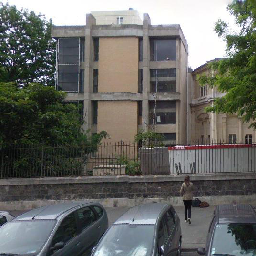}
    \caption*{GT}
  \end{subfigure}
  \begin{subfigure}[t]{0.12\linewidth}
    \includegraphics[width=\textwidth]{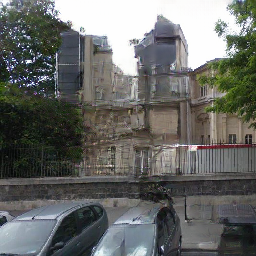}
    \caption*{ICT1}
  \end{subfigure}
  \begin{subfigure}[t]{0.12\linewidth}
    \includegraphics[width=\textwidth]{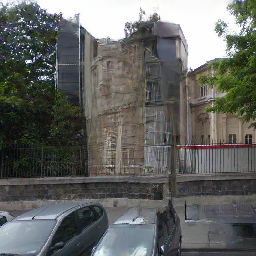}
    \caption*{ICT2}
  \end{subfigure}
  \begin{subfigure}[t]{0.12\linewidth}
    \includegraphics[width=\textwidth]{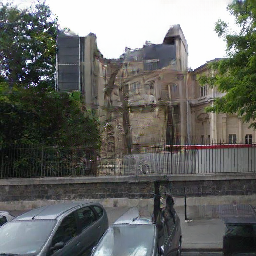}
    \caption*{ICT3}
  \end{subfigure}
  \begin{subfigure}[t]{0.12\linewidth}
    \includegraphics[width=\textwidth]{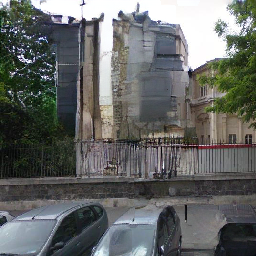}
    \caption*{LDM1}
  \end{subfigure}
  \begin{subfigure}[t]{0.12\linewidth}
    \includegraphics[width=\textwidth]{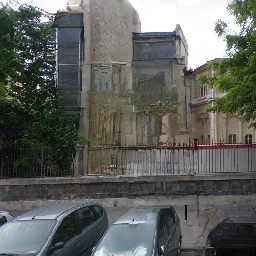}
    \caption*{LDM2}
  \end{subfigure}
  \begin{subfigure}[t]{0.12\linewidth}
    \includegraphics[width=\textwidth]{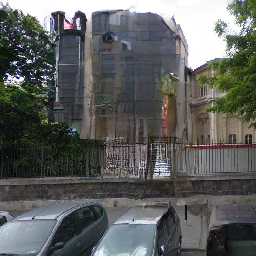}
    \caption*{LDM3}
  \end{subfigure}

  \smallskip
  \begin{subfigure}[t]{0.12\linewidth}
    \includegraphics[width=\textwidth]{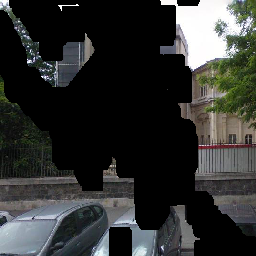}
    \caption*{Input}
  \end{subfigure}
  \begin{subfigure}[t]{0.12\linewidth}
    \includegraphics[width=\textwidth]{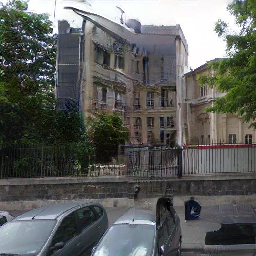}
    \caption*{MAT1}
  \end{subfigure}
  \begin{subfigure}[t]{0.12\linewidth}
    \includegraphics[width=\textwidth]{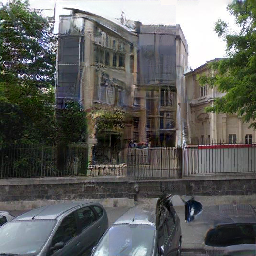}
    \caption*{MAT2}
  \end{subfigure}
  \begin{subfigure}[t]{0.12\linewidth}
    \includegraphics[width=\textwidth]{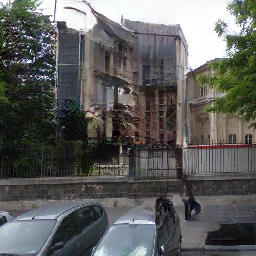}
    \caption*{MAT3}
  \end{subfigure}
  \begin{subfigure}[t]{0.12\linewidth}
    \includegraphics[width=\textwidth]{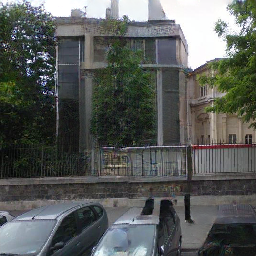}
    \caption*{Ours1}
  \end{subfigure}
  \begin{subfigure}[t]{0.12\linewidth}
    \includegraphics[width=\textwidth]{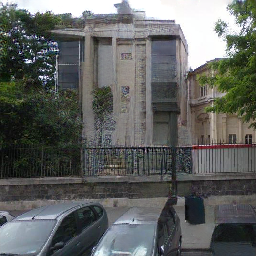}
    \caption*{Ours2}
  \end{subfigure}
  \begin{subfigure}[t]{0.12\linewidth}
    \includegraphics[width=\textwidth]{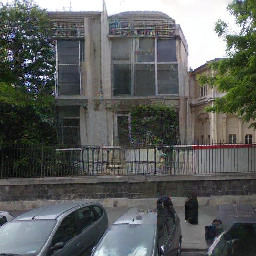}
    \caption*{Ours3}
  \end{subfigure}

  \smallskip
  \begin{subfigure}[t]{0.12\linewidth}
    \includegraphics[width=\textwidth]{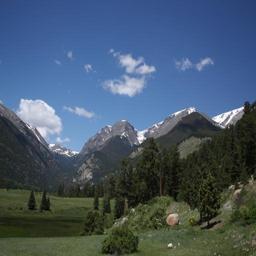}
    \caption*{GT}
  \end{subfigure}
  \begin{subfigure}[t]{0.12\linewidth}
    \includegraphics[width=\textwidth]{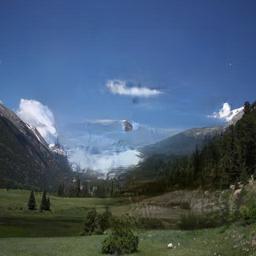}
    \caption*{ICT1}
  \end{subfigure}
  \begin{subfigure}[t]{0.12\linewidth}
    \includegraphics[width=\textwidth]{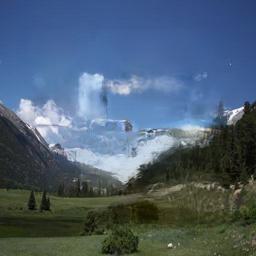}
    \caption*{ICT2}
  \end{subfigure}
  \begin{subfigure}[t]{0.12\linewidth}
    \includegraphics[width=\textwidth]{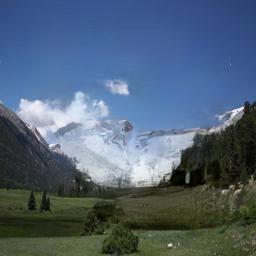}
    \caption*{ICT3}
  \end{subfigure}
  \begin{subfigure}[t]{0.12\linewidth}
    \includegraphics[width=\textwidth]{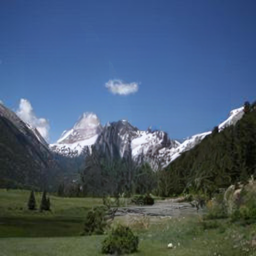}
    \caption*{LDM1}
  \end{subfigure}
  \begin{subfigure}[t]{0.12\linewidth}
    \includegraphics[width=\textwidth]{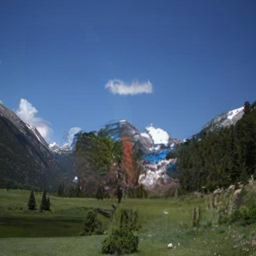}
    \caption*{LDM2}
  \end{subfigure}
  \begin{subfigure}[t]{0.12\linewidth}
    \includegraphics[width=\textwidth]{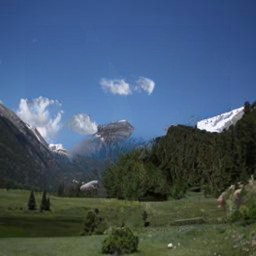}
    \caption*{LDM3}
  \end{subfigure}
  
  \smallskip
  \begin{subfigure}[t]{0.12\linewidth}
    \includegraphics[width=\textwidth]{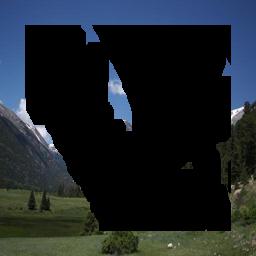}
    \caption*{Input}
  \end{subfigure}
  \begin{subfigure}[t]{0.12\linewidth}
    \includegraphics[width=\textwidth]{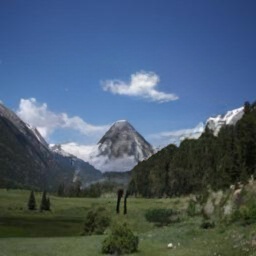}
    \caption*{MAT1}
  \end{subfigure}
  \begin{subfigure}[t]{0.12\linewidth}
    \includegraphics[width=\textwidth]{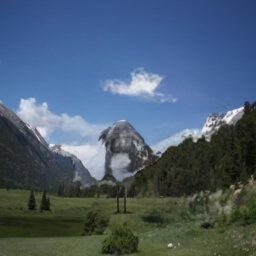}
    \caption*{MAT2}
  \end{subfigure}
  \begin{subfigure}[t]{0.12\linewidth}
    \includegraphics[width=\textwidth]{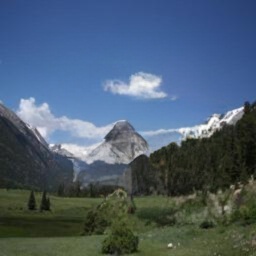}
    \caption*{MAT3}
  \end{subfigure}
  \begin{subfigure}[t]{0.12\linewidth}
    \includegraphics[width=\textwidth]{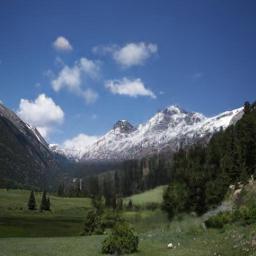}
    \caption*{Ours1}
  \end{subfigure}
  \begin{subfigure}[t]{0.12\linewidth}
    \includegraphics[width=\textwidth]{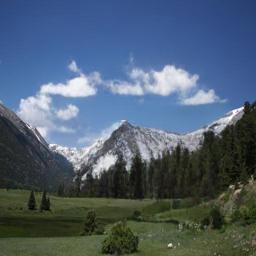}
    \caption*{Ours2}
  \end{subfigure}
  \begin{subfigure}[t]{0.12\linewidth}
    \includegraphics[width=\textwidth]{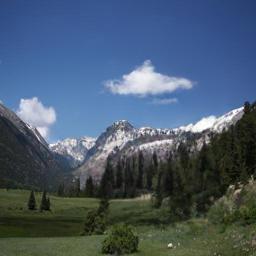}
    \caption*{Ours3}
  \end{subfigure}
  \caption{Visual comparison of diverse inpainting results  in Places and Paris Street View.}
  \label{fig:main_inp_result}
\end{figure*}

\begin{figure}[tb]
  \centering
  \begin{subfigure}{0.9\linewidth}
    \centering
    \includegraphics[width=\textwidth]{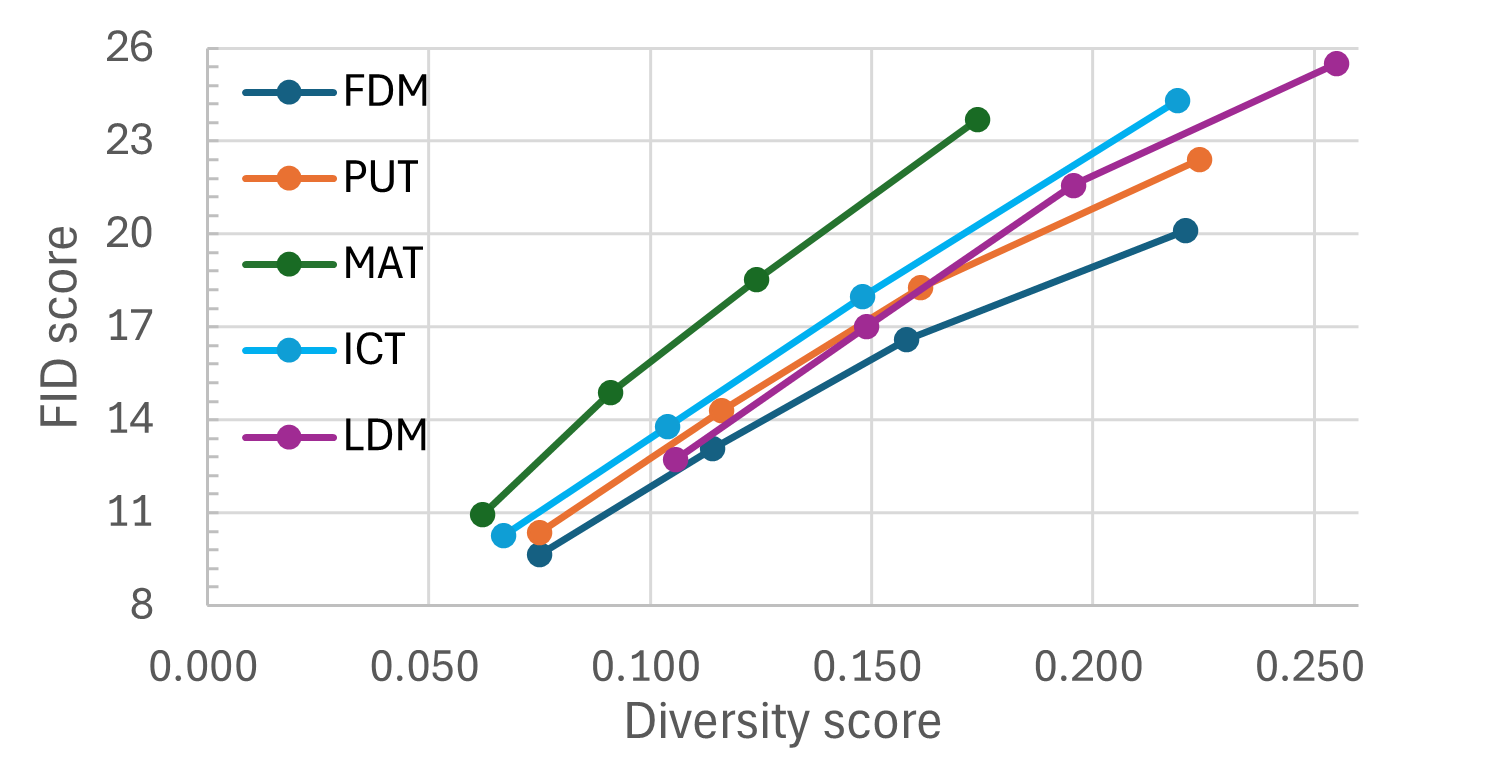}
    \caption{Paris Street View}
  \end{subfigure}
  \\
  \begin{subfigure}{0.9\linewidth}
    \centering
    \includegraphics[width=\textwidth]{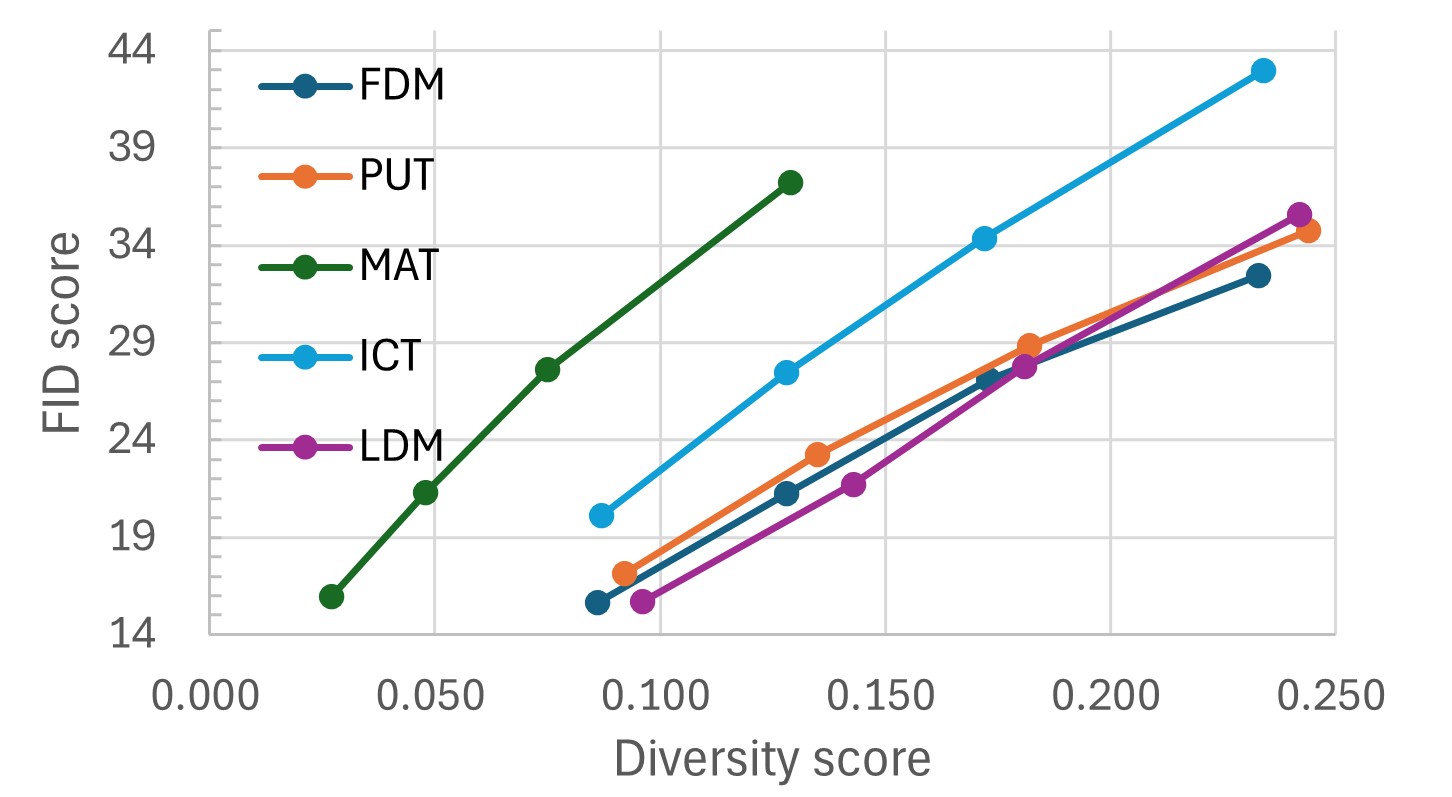}
    \caption{Places}
  \end{subfigure}
  \caption{Diversity score on different methods.}
  \label{fig:diversity_graph}
\end{figure}

\minisection{Qualitative comparisons}
Figure~\ref{fig:detail_inp_result} provides a detailed comparison between \putModel and FDM.
Results from \putModel contain color discrepancies and distorted structures.
For example, in the top-left sample, the region generated by \putModel appears brighter than the background, making the masked area clearly visible.
However, after applying feature dequantization through FDM, the generated color matches the background more accurately.
Another example is in the top-right sample, where the parking lines generated by \putModel should be straight but appear distorted.
Additionally, there are areas where asphalt is generated in grass colors.
In our results, parking lines are straight, and asphalt is represented in the same color as the background.

The pluralistic results of various methods are depicted in Figure~\ref{fig:main_inp_result}.
In this comparison, ICT produces significant artifacts, resulting in unnatural image.
While MAT is more natural than ICT, it lacks diversity and exhibits similar structures across images.
In contrast, our proposed method reliably generates diverse samples based on the probabilistic patch sampling of PUT.

\begin{table*}[tb]
  \caption{Quantitative results of different methods. \red{Red} indicates the best score, while \blue{blue} indicates the second-best score. Note that results for MAT and LDM on Places were obtained by training with 8.4 M and 1.8 M data, respectively. In contrast, the results of ICT, PUT, and our proposed method were obtained from training with the 0.24 M data same as \cite{wan2021high}.}
  \label{tab:main}
  \centering
  {\small
  \begin{tabular}{@{\extracolsep{4pt}}p{0.15\textwidth}rrrrrrrr@{}}
    \toprule[1.2pt]
    \multirow{3}{*}{Methods} & \multicolumn{4}{c}{Paris Street View} & \multicolumn{4}{c}{Places}\\
    \cline{2-5} \cline{6-9}
    & \multicolumn{2}{c}{FID} & \multicolumn{2}{c}{LPIPS} & \multicolumn{2}{c}{FID} & \multicolumn{2}{c}{LPIPS} \\
    \cline{2-3} \cline{4-5} \cline{6-7} \cline{8-9}
    & small & large & small & large & small & large & small & large \\
    \hline
    \ict & \blue{12.02} & 21.14 & 0.143 & 0.252 & 23.74 & 38.65 & 0.155 & 0.260 \\
    \mat & 13.22 & 21.45 & 0.151 & 0.266 & \red{18.39} & 32.09 & \blue{0.128} & \red{0.222} \\
    \ldm & 14.87 & 23.53 & 0.150 & 0.255 & 18.71 & \blue{31.66} & 0.136 & 0.265\\
    \putModel & 12.58 & \blue{20.52} & \blue{0.135} & \blue{0.240} & 20.25 & 32.01 & 0.137 & 0.240\\
    \ourPUT & \red{11.63} & \red{18.66} & \red{0.131} & \red{0.234} & \blue{18.46} & \red{29.67} & $\red{0.127}$ & \blue{0.230}\\
    \bottomrule[1.2pt]
  \end{tabular}}
\end{table*}

\minisection{Quantitative comparisons}
Table~\ref{tab:main} presents a comparison of methods across each evaluation metric.
In most cases, the proposed method achieved the best score across both datasets.
Particularly, our proposed method achieved the best FID score, especially with large masks.

The proposed method consistently outperforms \putModel, demonstrating the effectiveness of \ourmodule.
Moreover, performance generally ranks on the order of \ourmodule, PUT, and ICT, underscoring the importance of minimizing information loss to achieve superior results.
The significant performance improvement observed with large masks compared to small masks further highlights this point.
This is because a larger mask reduces the background information available for reference, thereby increasing the model's dependence on the provided information for accurate representation.

Figure~\ref{fig:diversity_graph} shows the diversity score of each method.
In the graph, the y-axis represents the FID score, and the x-axis represents the diversity score.
Each point corresponds to different mask ratios, ranging from 0.2 to 0.6, with intervals of 0.1 units.
An ideal position on the graph is in the bottom-right corner, indicating low FID scores and high diversity scores.
This indicates the ability to generate diverse and natural-looking images.
MAT, while demonstrating FID performance comparable to our proposed method on the Places dataset, shows a very low diversity score.
LDM has a high diversity score, but at higher mask ratios, it shows a lower FID score compared to our method.
Our proposed method achieved the best FID scores in most cases while still maintaining diversity of \putModel.

\subsection{Analysis}

\begin{table}[tb]
    \caption{Computational overhead of FDM on the Paris Street View dataset with a mask ratio of 20-30\%.}
    \label{tab:overhead}
    \centering
    {\small
    \begin{tabular}{p{0.138\textwidth}rrrr}
        \toprule[1.2pt]
        \multirow{2}{*}{Methods} & \multirow{2}{*}{$\sharp$ Param.} & \makecell[c]{Traning} & \multicolumn{2}{c}{Inference} \\
        \cline{4-5}
        & & \makecell[c]{Time} & \makecell[c]{Time} & \makecell[c]{FLOPs}\\
        \hline
        \putModel & 119M & 05d 09h & 25.236s & 59.331T \\
        \ourmodule & 3M & 00d 17h & 0.004s & 0.004T \\ \hline\hline
        \ourPUT & 122M & 06d 02h & 25.240s & 59.335T \\
        \toprule[1.2pt]
    \end{tabular}}
\end{table}

\minisection{Computational overhead of FDM}
Table~\ref{tab:overhead} shows the computational overhead of FDM in training and inference times.
During training, it occupies only 12\% of the total training time.
Training FDM with sampling, estimated based on the required time for one iteration, would take approximately 131 days.
To tackle this problem, we suggested training FDM using ground truth without sampling.
With our training strategy, FDM can be trained in just 17 hours, which is 184 times faster than the naive training method.
Therefore, we propose a method to practically apply FDM to inpainting through ground-truth training.
The number of parameters for FDM is approximately 2.5\% of the total number of parameters.
The FLOPs for FDM is approximately 0.01\% of the total FLOPs.
Moreover, the inference time accounts for only about 0.02\% of the overall time.
The result demonstrates that FDM improves generation quality with a minimal time overhead.

\begin{figure}[tb]
  \centering
  \begin{subfigure}{0.49\linewidth}
    \centering
    \includegraphics[width=\textwidth]{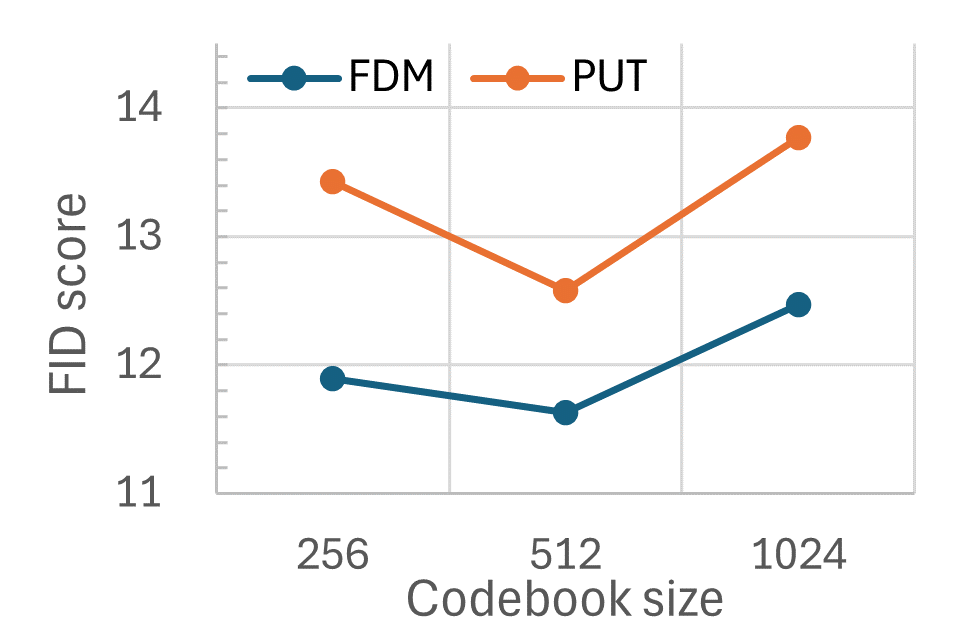}
    \caption{Small mask}
  \end{subfigure}
  \begin{subfigure}{0.49\linewidth}
    \centering
    \includegraphics[width=\textwidth]{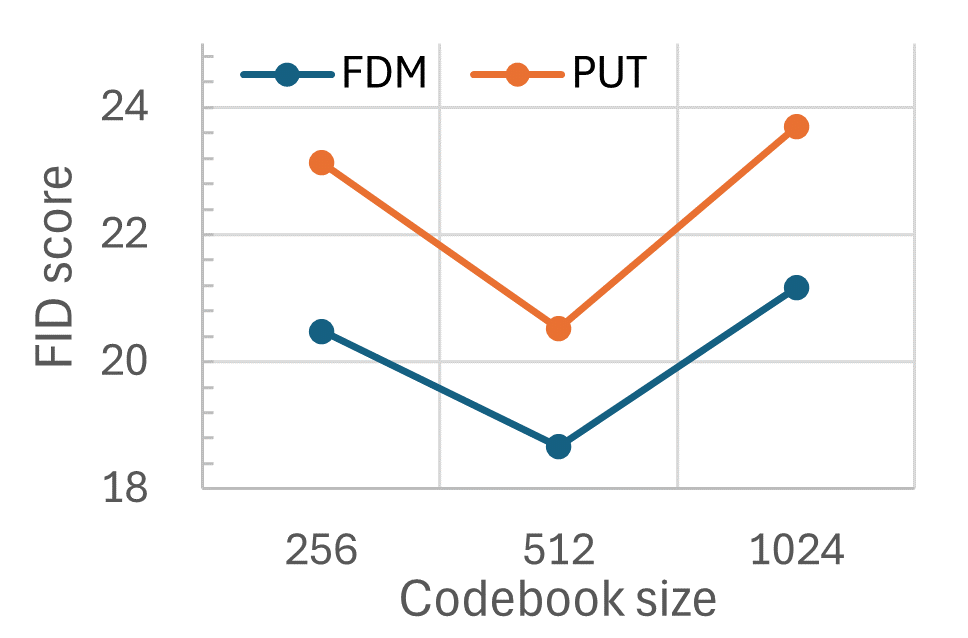}
    \caption{Large mask}
  \end{subfigure}
  \caption{FID scores based on codebook size in the Paris Street View
dataset.}
  \label{fig:codebook}
\end{figure}

\minisection{Relationship between codebook size and \ourmodule's effectiveness}
Figure ~\ref{fig:codebook} is the FID score graph based on codebook size.
Inpainting performance does not always scale proportionally with codebook size, unlike reconstruction performance.
Therefore, increasing codebook size is less effective than FDM in addressing information loss.
Moreover, FDM consistently improves performance regardless of codebook size.
Given its consistent performance improvements and simplicity, FDM provides an efficient solution for addressing information loss.

\begin{table}[tb]
    \caption{L2 distance between generated feature and ideal feature.}
    \label{tab:rec_distance}
    \centering
    {\small
    \begin{tabular}{lrrrr}
        \toprule[1.2pt]
        Mask ratio & 0.2-0.3 & 0.3-0.4 & 0.4-0.5 & 0.5-0.6 \\
        \hline
        PUT & 3.685& 3.675& 3.678& 3.669 \\
        \ourPUT & 2.873& 2.876& 2.893 & 2.901 \\ \hline\hline
        Difference & 0.812& 0.799& 0.785& 0.768 \\
        \toprule[1.2pt]
    \end{tabular}}
\end{table}

\minisection{Feature dequantization}
Table~\ref{tab:rec_distance} presents L2 distance between the generated features and the corresponding ideal features in the Places dataset.
The ideal features are obtained by encoding the original image using the encoder.
The results indicate that the features restored by FDM closely align with the original feature space across all mask ratios.
Our proposed method generates features that are closer to the ideal features compared to those generated by \putModel, resulting in more plausible results.

\subsection{Applying FDM to Image Generation Tasks}
\label{sec:img_gen}

\begin{figure}[tb]
  \centering
  {\small
  \begin{subfigure}[t]{0.05\linewidth}
    \raisebox{2.40\linewidth}{\rotatebox[origin=t]{90}{\vqgan}}\\
    \raisebox{-2.40\linewidth}{\rotatebox[origin=t]{90}{Ours}}
  \end{subfigure}}
  \begin{subfigure}[t]{0.30\linewidth}
    \includegraphics[width=\textwidth]{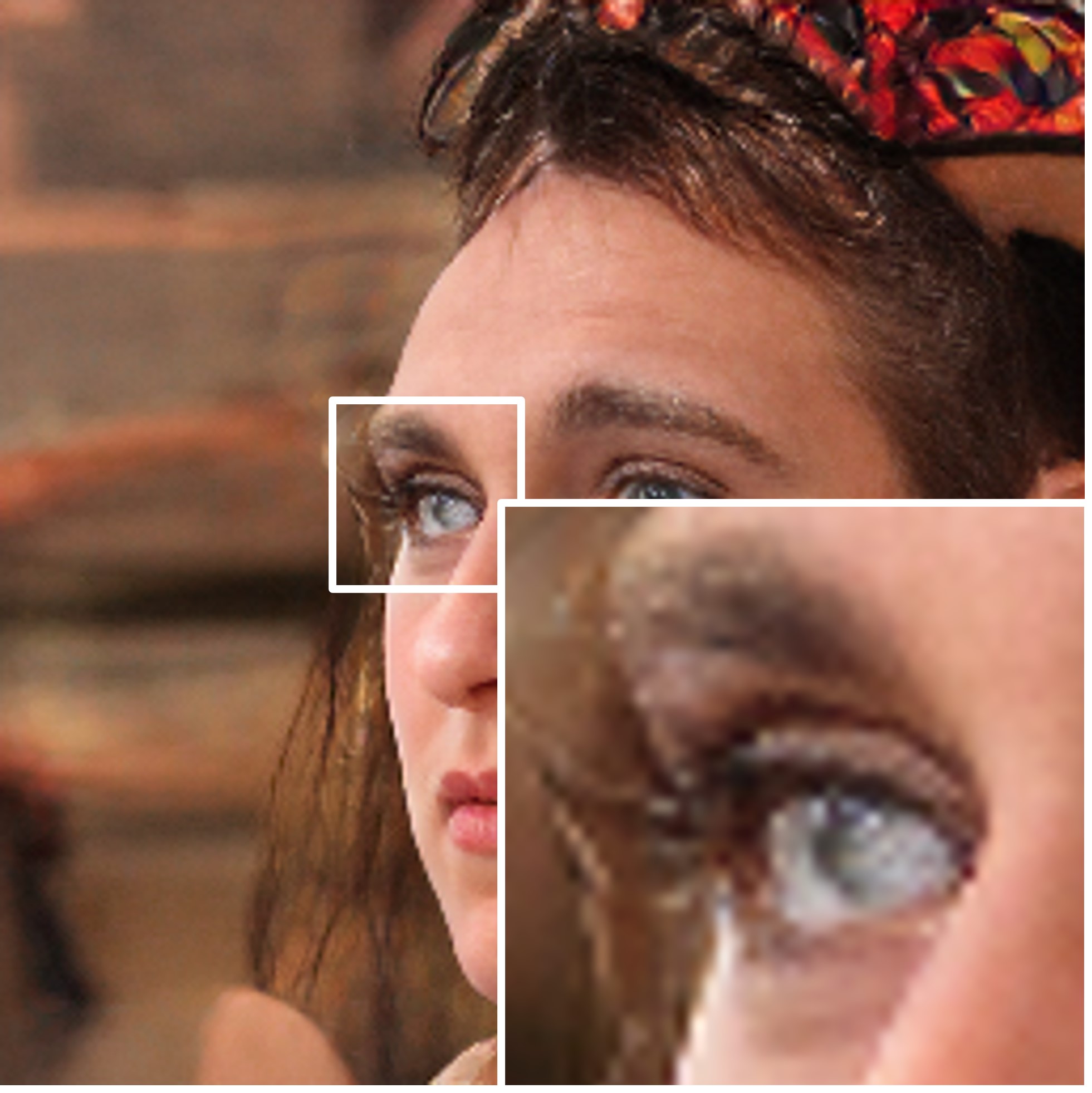} \\
    \includegraphics[width=\textwidth]{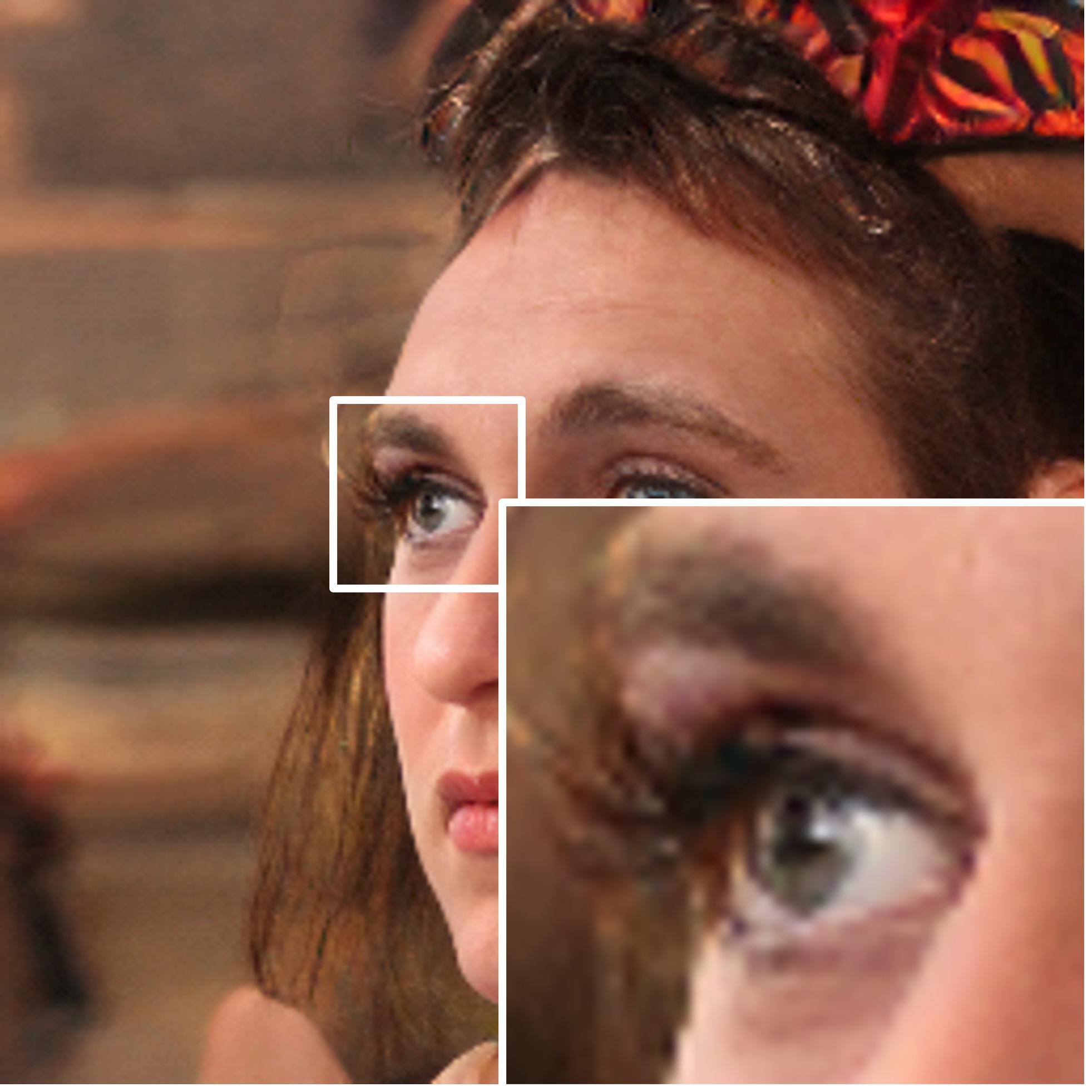}
    \caption*{FFHQ}
  \end{subfigure}
  \begin{subfigure}[t]{0.30\linewidth}
    \includegraphics[width=\textwidth]{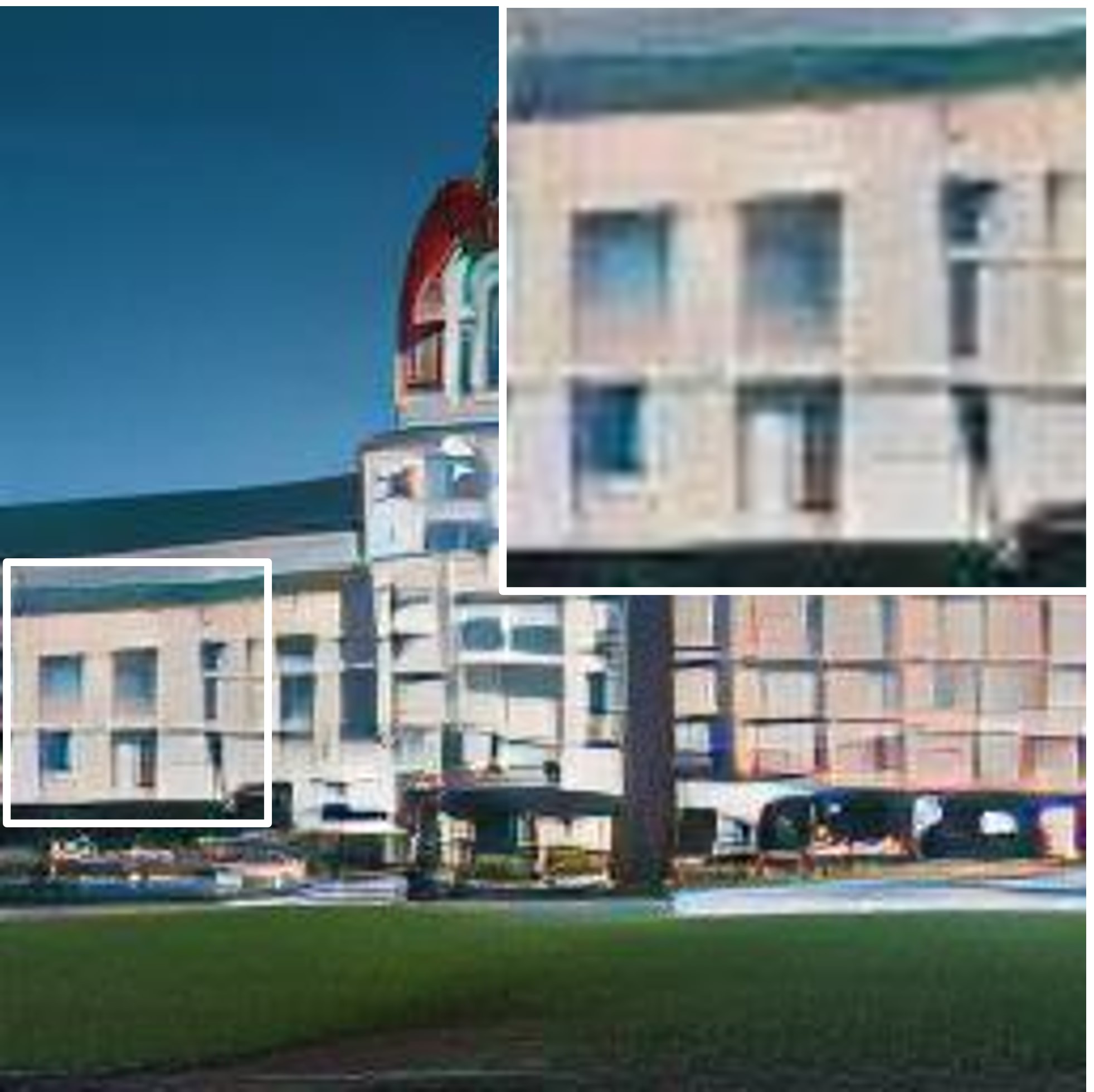} \\
    \includegraphics[width=\textwidth]{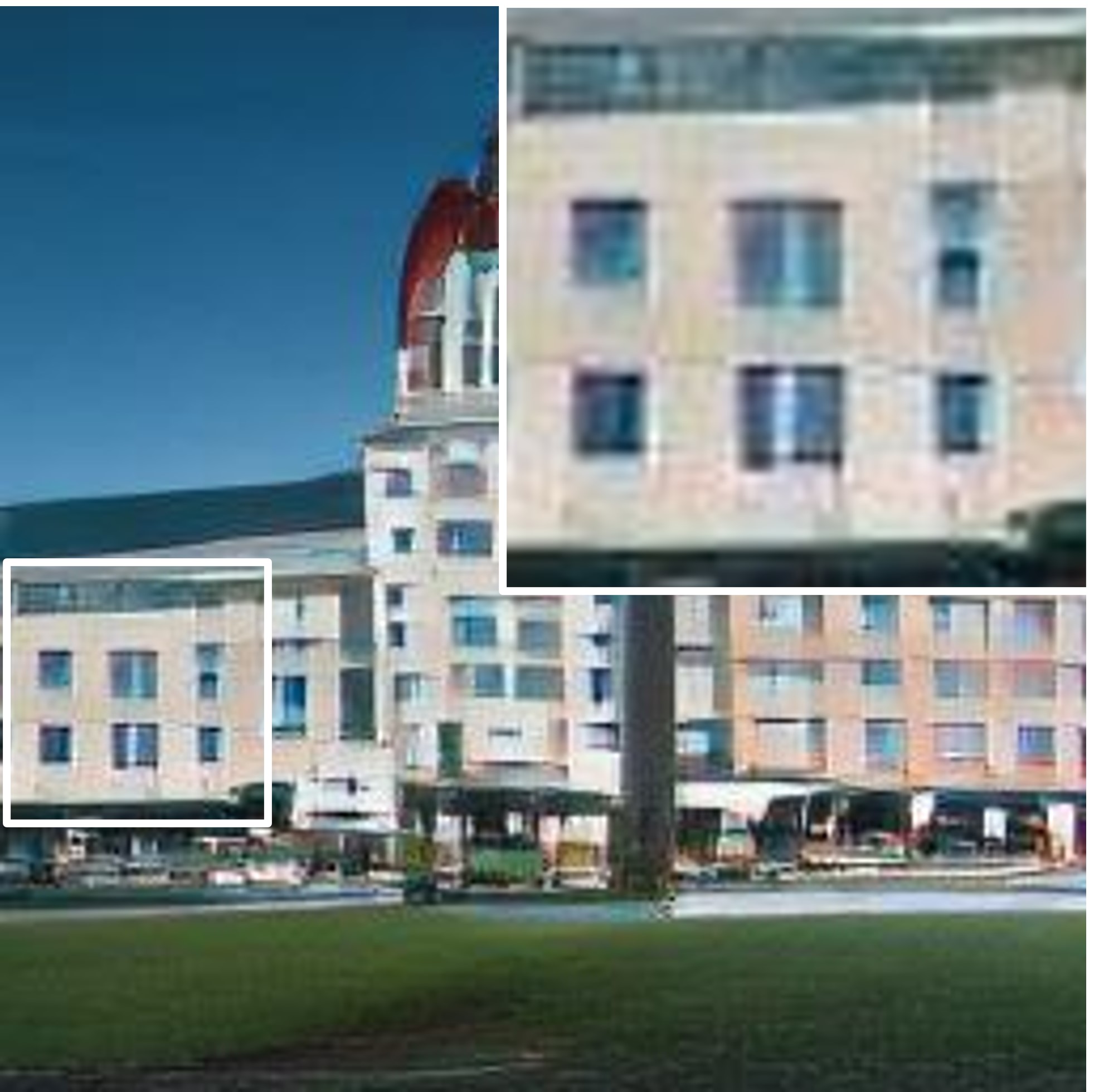} 
    \caption*{ADE20K}
  \end{subfigure}
  \begin{subfigure}[t]{0.30\linewidth}
    \includegraphics[width=\textwidth]{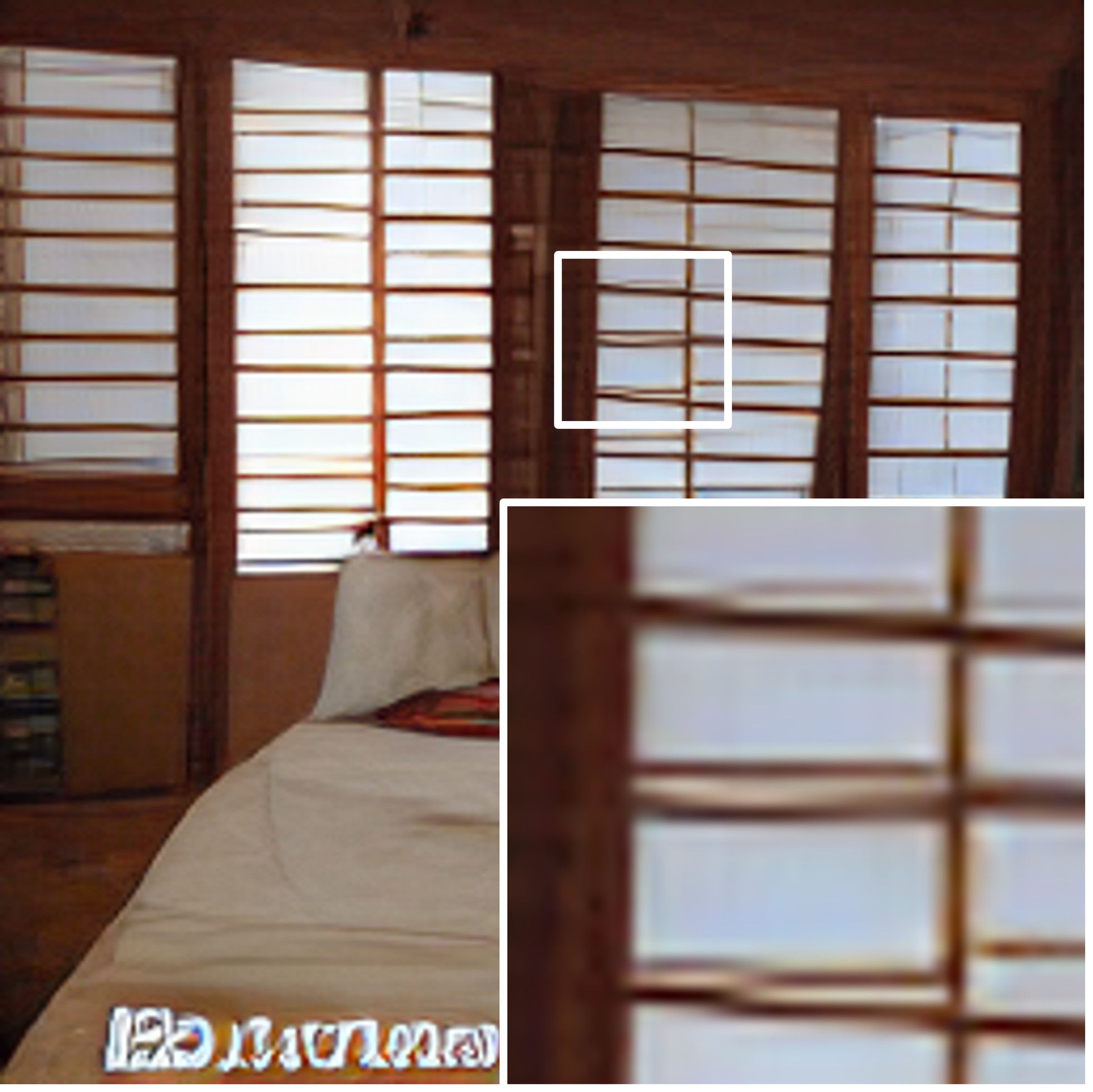} \\
    \includegraphics[width=\textwidth]{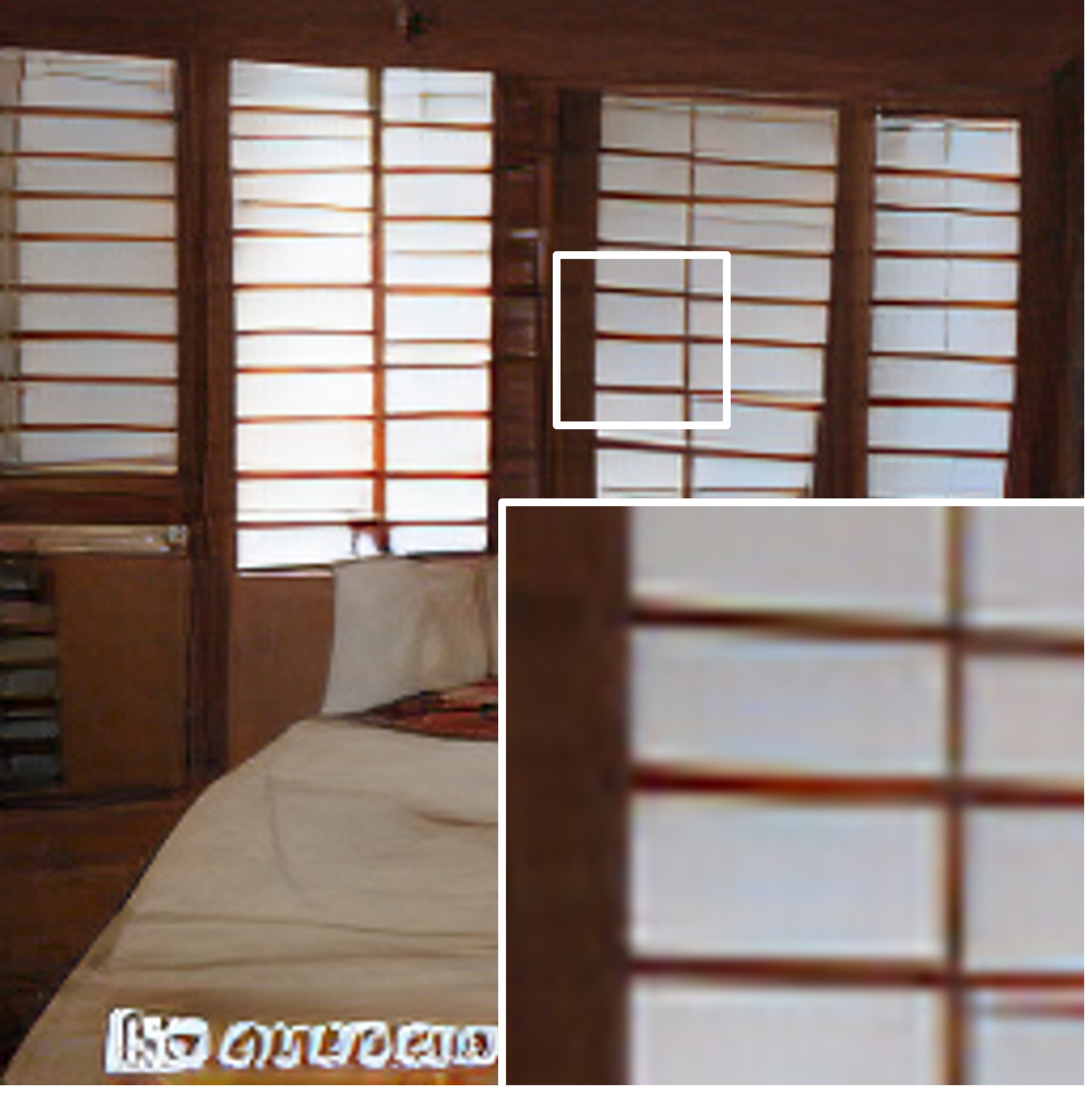}
    \caption*{ImageNet}
  \end{subfigure}
  \caption{Detail comparison between ours and \vqgan.}
  \label{fig:img_gen}
\end{figure}

\begin{table}[tb]
  \caption{FID score comparison for various image generation tasks.}
  \label{tab:img_gen}
  \centering
  {\small
  \begin{tabular}{@{\extracolsep{1pt}}p{0.17\textwidth}ccc@{}}
    \toprule[1.2pt]
    \multirow{2}{*}{Methods} & Uncond. & Semantic & Class \\
    & (FFHQ) & (ADE20K) & (ImageNet)\\
    \hline
    \vqgan & 17.03 & 33.26 & 14.65 \\
    \ourVQGAN & $\mathbf{15.84}$ & $\mathbf{31.32}$ & $\mathbf{13.46}$\\
    \bottomrule[1.2pt]
  \end{tabular}}
\end{table}

We conducted additional experiments to demonstrate that \ourmodule can improve \vqgan in various image generation tasks: unconditional image generation, semantic-conditional image generation, and class-conditional image synthesis.
For the three tasks, we use the FFHQ\cite{karras2019style}, ADE20K\cite{zhou2017scene}, and ImageNet\cite{deng2009imagenet} datasets, respectively.

\minisection{Experimental settings}
The experiments are conducted by adding FDM to \vqgan\cite{esser2021taming}, a prominent vector-quantization based image generation model.
FDM is applied after all patches are predicted, as described in Section ~\ref{sec:architecture}.
For unconditional and class-conditional, FDM takes only the quantized feature  $\quantInpFeat$ as input without a mask.
The training procedure follows Section ~\ref{sec:training_procedure}, and Phase 1 is skipped since we use the pre-trained model provided by \cite{esser2021taming}.
We follow the training settings provided by \cite{esser2021taming}.

\minisection{Qualitative comparisons}
Figure ~\ref{fig:img_gen} provides a detailed comparison between \vqgan and Ours.
Images generated by \vqgan often lack detailed representation.
For instance, in the ADE20K sample at the center, \vqgan-produced images suffer from blurring and gradient effects, which obscure the distinction between building windows and walls.
However, in the images generated using our method, the window frames are clear and distinguishable from the walls.
Another example from the rightmost sample shows that window frames in \vqgan-generated images are split into two sections, whereas our method produces window frames as single, continuous pieces.

\minisection{Quantitative comparisons}
Table~\ref{tab:img_gen} shows the FID score comparison for image synthesis.
After applying \ourmodule, there was an improvement in FID scores across all tasks.
This demonstrates that our method can effectively and simply enhance vector-quantization based models across various image generation tasks, not just inpainting. 
\section{Conclusion and Limitations}

In this paper, we studied the pluralistic image inpainting (PII) problem which offers multiple plausible solutions for missing image parts.
We introduced \ourmodule (Feature Dequantization Module), which enhances representational capacity through feature dequantization, thereby improving the details of generated images.
\ourmodule can be seamlessly applied during the inference phases with minimal overhead.
In addition, we proposed an efficient training method to train \ourmodule which dramatically reduces the training cost by removing the sampling in the training phase.
Furthermore, through experiments, our proposed method has demonstrated effectiveness  across various image generation tasks, not just limited to image inpainting.

In \vqgan-based PII, the sequence in which patches are inpainted plays an important role in defining both the quality and variety of the output images.
However, since our method \ourmodule is applied after the feature-sampler, it cannot affect the order of inpainting.
To address this, we can consider configuring \ourmodule to perform dequantization on the unmasked parts of the input containing masked patches.
This enables \ourmodule to be applied during the sampling process, allowing it to affect the inpainting order.
\section{Acknowledgments}

This research was supported by the MSIT(Ministry of Science and ICT), Korea, under the Convergence security core talent training business support program(IITP-2024-RS-2024-00423071) supervised by the IITP(Institute of Information \& Communications Technology Planning \& Evaluation).
This work was supported by Institute of Information \& communications Technology Planning \& Evaluation(IITP) grant funded by the Korea government(MSIT) (No.RS-2023-00261068, Development of Lightweight Multimodal Anti-Phishing Models and Split-Learning Techniques for Privacy-Preserving Anti-Phishing)


{\small
\bibliographystyle{ieee_fullname}
\bibliography{egbib}
}

\end{document}